\documentclass{article}
    \PassOptionsToPackage{numbers, compress}{natbib}

    \usepackage[final]{neurips_data_2023}

\usepackage[utf8]{inputenc} 
\usepackage[T1]{fontenc}    
\usepackage{hyperref}       
\usepackage{url}            
\usepackage{booktabs}       
\usepackage{amsfonts}       
\usepackage{nicefrac}       
\usepackage{microtype}      
\usepackage[dvipsnames]{xcolor}         
\usepackage{graphics,graphicx,caption,float,subcaption,booktabs,xcolor,multirow,array,color,ifthen,tabu,colortbl,dblfloatfix,url,xparse,mathtools,patchcmd,algorithm,algorithmic,amssymb,xspace,nicefrac,microtype,amsmath,amsthm,amsfonts,bm,ragged2e,tikz,stackengine,etoolbox,xpatch,enumerate,xstring,setspace,tabularx,makecell,changepage,cuted,titlesec,enumitem,wrapfig,tcolorbox, threeparttable}
\usepackage{pifont}
\newcommand{\cmark}{\textcolor{ForestGreen}{\ding{51}}}%
\newcommand{\xmark}{\textcolor{BrickRed}{\ding{55}}}%

\def\1{mathbb{1}}

\def\0{{\bf 0}}
\def\1{{\bf 1}}

\definecolor{purple}{rgb}{0.56,0.27,0.68}
\definecolor{red}{rgb}{0.95,0.4,0.4}
\definecolor{purered}{rgb}{1,0,0}
\definecolor{blue}{rgb}{0.4,0.4,0.95}
\definecolor{darkblue}{rgb}{0,0,0.8}
\definecolor{grey}{rgb}{0.6,0.6,0.6}
\definecolor{col1}{RGB}{232, 161, 148}
\definecolor{col2}{RGB}{148, 187, 232}
\definecolor{col3}{RGB}{206, 239, 255}
\definecolor{lightgrey}{rgb}{0.85,0.85,0.85}
\definecolor{lightlightgrey}{rgb}{0.9,0.9,0.9}
\definecolor{verylightBG}{rgb}{0.9,0.99,0.99}
\definecolor{darkgreen}{rgb}{0.3, 0.75, 0.3}

\graphicspath{{./figures/}}   

\newcommand\note[1]{\textcolor{black}{#1}}

\title{A High-Resolution Dataset for Instance Detection with Multi-View Instance Capture}

\author{%
  Qianqian Shen$^{1, }$\thanks{co-first authors.} \ \ Yunhan Zhao$^{2, *}$ \ \ Nahyun Kwon$^3$ \ \ Jeeeun Kim$^{3}$ \ \ Yanan Li$^{1}$  \ \ Shu Kong$^{3,4}$ 
  \\
  $^1$Zhejiang Lab \quad\quad $^2$UC-Irvine \quad\quad $^3$Texas A\&M University  \quad\quad $^4$University of Macau \\ \\
  \href{https://github.com/insdet/instance-detection}{\em Dataset and open-source code}
}

\begin{document}

\maketitle

\begin{abstract}
Instance detection (InsDet) is a long-lasting problem in robotics and computer vision, aiming to detect object instances (predefined by some visual examples) in a cluttered scene. Despite its practical significance, its advancement is overshadowed by Object Detection, which aims to detect objects belonging to some predefined classes. One major reason is that current InsDet datasets are too small in scale by today's standards. For example, the popular InsDet dataset GMU (published in 2016) has only 23 instances, far less than COCO (80 classes), a well-known object detection dataset published in 2014. We are motivated to introduce a new InsDet dataset and protocol. First, we define a realistic setup for InsDet: training data consists of multi-view instance captures, along with diverse scene images allowing synthesizing training images by pasting instance images on them with free box annotations. Second, we release a real-world database, which contains multi-view capture of 100 object instances, and high-resolution (6k$\times$8k) testing images. Third, we extensively study baseline methods for InsDet on our dataset, analyze their performance and suggest future work. Somewhat surprisingly, using the off-the-shelf class-agnostic segmentation model (Segment Anything Model, SAM) and the self-supervised feature representation DINOv2 performs the best, achieving $>$10 AP better than end-to-end trained InsDet models that repurpose object detectors (e.g., FasterRCNN and RetinaNet).
\end{abstract}

\section{Introduction}
\label{sec:intro}

Instance detection (InsDet) requires detecting specific object instances (defined by some visual examples) from a scene image~\cite{dwibedi2017cut}. 
It is practically important in robotics, e.g., elderly-assistant robots need to fetch specific items (\textit{my}-cup vs. \textit{your}-cup) from a cluttered kitchen~\cite{savage2022robots}, micro-fulfillment robots for the retail need to pick items from mixed boxes or shelves~\cite{bormann2021real}.

\begin{figure*}[t]
\begin{center}
\includegraphics[width=1\linewidth]{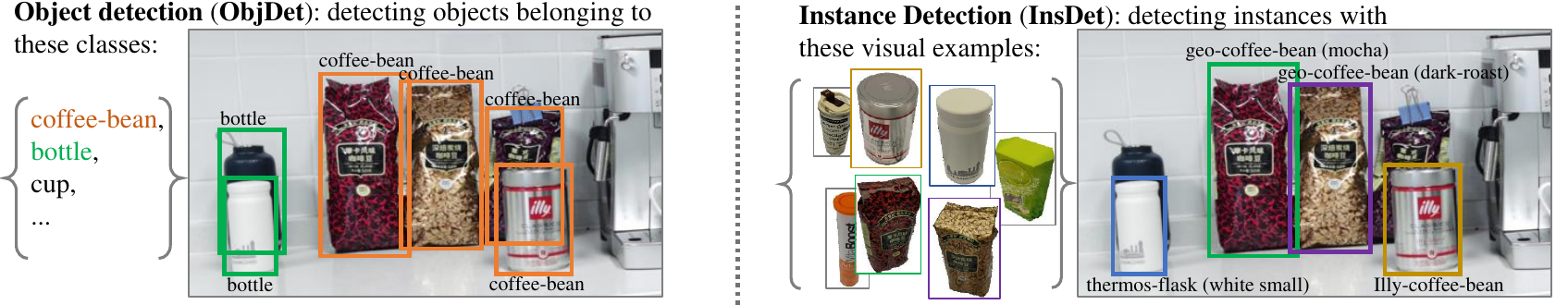}
\end{center}
\vspace{-4mm}
\caption{\small
\note{
{\bf Object detection (ObjDet) vs. instance detection (InsDet)}.
ObjDet aims to detect all objects belonging to some predefined classes, whereas InsDet requires detecting specific object instances defined by some visual examples.
Loosely speaking, InsDet treats a single object instance as a class compared to ObjDet. 
Please refer to Fig.~\ref{fig:data-setup}-right for the challenge of InsDet, which is the focus of our work.
}
}
\label{fig:ObjDet-vs-InsDet}
\vspace{-4.5mm}
\end{figure*}

{\bf Motivation}.
\note{
InsDet receives much less attention than the related problem of Object Detection (ObjDet), which aims to detect all objects belonging to some predefined classes~\cite{Lin2014MicrosoftCC, NIPS2015_fasterRCNN, lin2017focal, zhang2022dino}. Fig.~\ref{fig:ObjDet-vs-InsDet} compares the two problems.
\emph{One major reason is that there are not large-enough InsDet datasets by today's standards.} For example, the popular InsDet dataset GMU (published in 2016)~\cite{Georgakis2016MultiviewRD} has only 23 object instances while the popular ObjDet dataset COCO has 80 object classes (published in 2014)~\cite{Lin2014MicrosoftCC}. 
Moreover, \emph{there are no unified protocols in the literature of InsDet.} 
The current InsDet literature mixes multiple datasets to simulate training images and testing scenarios~\cite{dwibedi2017cut}.
Note that the training protocol of InsDet does not follow that of ObjDet, which has training images annotated with bounding boxes.
Differently, for InsDet,\footnote{In real-world applications (e.g., robot learning), it is infeasible to place objects in diverse scenes, take scene photos, then annotate instances using boxes towards training images (cf. training data in object detection).} its setup should have profile images of instances (cf. right in Fig.~\ref{fig:ObjDet-vs-InsDet}) and optionally diverse background images not containing such instances~\cite{dwibedi2017cut}. 
We release a new dataset and present a unified protocol to foster the InsDet research.
}

\begin{figure*}[t]
\begin{center}
\includegraphics[width=1\linewidth]{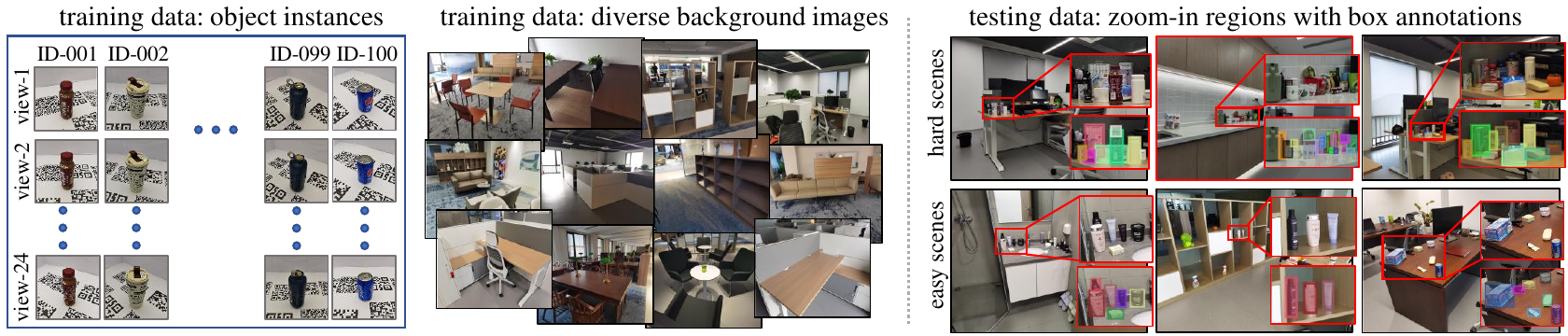}
\end{center}
\vspace{-4mm}
\caption{\small
{\bf Overview of our instance detection dataset}.
{\bf Left}: It contains 100 distinct object instances. For each of them, we capture 24 profile photos from multiple views. We paste QR code images beneath objects to allow relative camera estimation (e.g., by COLMAP~\cite{schonberger2016structure}), just like other existing datasets~\cite{Hinterstoier2012ModelBT, Brachmann2014Learning6O}.
{\bf Middle}: We take photos in random scenes (which do not contain any of the 100 instances) as background images. The background images can be optionally used to synthesize training data, e.g., pasting the foreground instances on them towards box-annotated training images~\cite{Lai2014UnsupervisedFL, dwibedi2017cut, georgakis2017synthesizing} as used in the object detection literature~\cite{Lin2014MicrosoftCC}.
{\bf Right}: high-resolution (6k$\times$8k) testing images of clutter scenes contain diverse instances, including some of the 100 predefined instances and other uninterested ones. The goal of InsDet is to detect the predefined instances in these testing images. 
From the zoom-in regions, we see the scene clutters make InsDet a rather challenging problem.
}
\vspace{-3mm}
\label{fig:data-setup}
\end{figure*}

{\bf Overview of our dataset} is presented in Fig.~\ref{fig:data-setup}.
In our dataset, profile images (3072x3072) of object instances and testing images (6144x8192) are high-resolution captured by a Leica camera (commonly used in today's cellphones). This inexpensive camera is deployable in current or future robot devices. Hence, our dataset simulates real-world scenarios, e.g., robotic navigation in indoor scenes.
Even with high-resolution images, objects in testing images appear small, taking only a tiny region in the high-res images.
This demonstrates a clear challenge of InsDet in our dataset.
Therefore, our dataset allows studying InsDet methods towards real-time operation on high-res (as future work).

{\bf Preview of technical insights}.
On our dataset, we revisit existing InsDet methods~\cite{Lai2014UnsupervisedFL, dwibedi2017cut, georgakis2017synthesizing}. 
Perhaps the only InsDet framework is cut-paste-learn~\cite{dwibedi2017cut}, which cuts instances from their profile images, pastes them on random background images (so being able to derive ``free'' bounding boxes annotations), and trains InsDet detectors on such data by following that of ObjDet (e.g., FasterRCNN~\cite{NIPS2015_fasterRCNN}). We study this framework, train different detectors, and confirm that the state-of-the-art transformer-based detector DINO~\cite{zhang2022dino} performs the best, achieving 27.99 AP, significantly better than CNN-based detector FasterRCNN (19.52 AP).
Further, we present a non-learned method that runs off-the-shelf proposal detectors (SAM~\cite{kirillov2023segment} in our work) to generate object proposals and use self-supervised learned features (DINO$_{f}$~\cite{caron2021emerging}\footnote{We add subscript $_{f}$ to indicate that DINO$_{f}$~\cite{caron2021emerging} is the self-supervised learned feature extractor; distinguishing it from a well-known object detector DINO~\cite{zhang2022dino}.} and DINOv2$_{f}$~\cite{oquab2023dinov2}) to find matched proposals to instances' profile images.
Surprisingly, this non-learned method resoundingly outperforms end-to-end learning methods, i.e., SAM+DINOv2$_{f}$ achieves 41.61 AP, much better than DINO (27.99 AP)~\cite{zhang2022dino}.

{\bf Contributions}. We make three major contributions.
\begin{enumerate}[noitemsep,  topsep=1pt]
    \item We formulate the InsDet problem with a unified protocol and release a challenging dataset consisting of both high-resolution profile images and high-res testing images. 

    \item We conduct extensive experiments on our dataset and benchmark representative methods following the cut-paste-learn framework~\cite{dwibedi2017cut}, showing that stronger detectors perform better.
    
    \item We present a non-learned method that uses an off-the-shelf proposal detector (i.e., SAM~\cite{kirillov2023segment}) to produce proposals, and self-supervised learned features (e.g., DINOv2$_f$~\cite{oquab2023dinov2}) to find instances (which are well matched to their profile images). This simple method significantly outperforms the end-to-end InsDet models.
\end{enumerate}

\section{Related Work}

{\bf Instance Detection (InsDet)} is a long-lasting problem in computer vision and robotics~\cite{zhou2019objects, dwibedi2017cut, mercier2021deep, ancha2019combining, georgakis2016multiview, hodavn2019photorealistic, bormann2021real}, referring to detecting specific object instances in a scene image. 
Traditional InsDet methods use keypoint matching~\cite{quadros2012occlusion} or template matching~\cite{hinterstoisser2011gradient}; more recent ones train deep neural networks to approach InsDet~\cite{mercier2021deep}.
Some others focus on obtaining more training samples by rendering realistic instance examples~\cite{kehl2017ssd, hodavn2019photorealistic}, data augmentation~\cite{dwibedi2017cut}, and synthesizing training images by cutting instances as foregrounds and pasting them to background images~\cite{Lai2014UnsupervisedFL, dwibedi2017cut, georgakis2017synthesizing}. 
Speaking of InsDet datasets, 
\cite{Georgakis2016MultiviewRD} collects scene images from 9 kitchen scenes with RGB-D cameras and defines 23 instances of interest to annotate with 2D boxes on scene images;
\cite{hodavn2019photorealistic} creates 3D models of 29 instances from 6 indoor scenes, and uses them to synthesize training and testing data;
\cite{bormann2021real} creates 3D mesh models of 100 grocery store objects, renders 80 views of images for each instance, and uses them to synthesize training data.

As for benchmarking protocol of InsDet,
\cite{dwibedi2017cut} synthesizes training data from BigBird~\cite{Singh2014BigBIRDAL} and UW Scenes~\cite{lai2011large} and tests on the GMU dataset~\cite{Georgakis2016MultiviewRD};
\cite{hodavn2019photorealistic} trains on their in-house data and test on LM-O~\cite{Brachmann2014Learning6O} and Rutgers APC~\cite{rennie2016dataset} datasets.
Moreover, some works require hardware-demanding setups~\cite{bormann2021real}, 
some synthesize both training and testing data~\cite{hodavn2019photorealistic, Lai2014UnsupervisedFL},
while others mix existing datasets for benchmarking~\cite{dwibedi2017cut}. 
Given that the modern literature on InsDet lacks a unified benchmarking protocol (till now!), we introduce a more realistic unified protocol along with our InsDet dataset, allowing fairly benchmarking methods and fostering research of InsDet.

{\bf Object Detection (ObjDet)} is a fundamental computer vision problem~\cite{felzenszwalb2009object, Lin2014MicrosoftCC, NIPS2015_fasterRCNN}, requiring detecting all objects belonging to some predefined categories. 
The prevalent ObjDet detectors adopt convolutional neural networks (CNNs) as a backbone and a detector-head for proposal detection and classification, typically using bounding box regression and a softmax-classifier.
Approaches can be grouped into two categories: one-stage detectors~\cite{redmon2016you, liu2016ssd, redmon2018yolov3, wang2022yolov7} and two-stage detectors~\cite{girshick2015fast,cai2018cascade}. 
One-stage detectors predict candidate detection proposals using bounding boxes and labels at regular spatial positions over feature maps; two-stage detectors first produce detection proposals, then perform classification and bounding box regression for each proposal. 
Recently, the transformer-based detectors transcend CNN-based detectors~\cite{carion2020end, zhu2020deformable, zhang2022dino}, yielding much better performance on various ObjDet benchmarks.
Different from ObjDet, InsDet requires distinguishing individual object instances within a class. Nevertheless, to approach InsDet, the common practice is to repurpose ObjDet detectors by treating unique instances as individual classes. 
We follow this practice and benchmark various ObjDet methods on our InsDet dataset.

{\bf Pretrained Models}.
Pretraining is an effective way to learn features from diverse data.
For example, training on the large-scale ImageNet dataset for image classification~\cite{deng2009imagenet}, a neural network can serve as a powerful feature extractor for various vision tasks~\cite{donahue2014decaf, sharif2014cnn}.
Object detectors trained on the COCO dataset~\cite{Lin2014MicrosoftCC} can serve as a backbone allowing finetuning on a target domain to improve detection performance~\cite{li2019analysis}.
Such pretraining requires human annotations which can be costly. Therefore, self-supervised pretraining has attracted increasing attention and achieved remarkable progress~\cite{chen2020simple, he2020momentum, caron2021emerging, oquab2023dinov2}.
Moreover, the recent literature shows that pretraining on much larger-scale data can serve as a foundation model for being able to perform well across domains and tasks. For example, the Segment Anything Model (SAM) pretrains a class-agnostic proposal detector on web-scale data and shows an impressive ability to detect and segment diverse objects in the wild~\cite{kirillov2023segment}.
In this work, with our high-res InsDet dataset, we explore a non-learned method by using publicly available pretrained models. We show that such a simple method significantly outperforms end-to-end learned InsDet detectors.

\section{Instance Detection: Protocol and Dataset}
\label{sec:insdet-dataset}

In this section, we formulate a realistic unified InsDet protocol and introduce the new dataset.
We release our dataset under the MIT License, hoping to contribute to the broader research community.

\subsection{The Protocol}

Our InsDet protocol is motivated by real-world indoor robotic applications. In particular, we consider the scenario that assistive robots must locate and recognize instances to fetch them in a cluttered indoor scene~\cite{savage2022robots}, where InsDet is a crucial component.
Realistically, for a given object instance, the robots should see it only from a few views (\emph{at the training stage}), and then accurately detect it \emph{in a distance} in \emph{any} scenes (\emph{at the testing stage}).
Therefore, we suggest the protocol specifying the training and testing setups below. We refer the readers to Fig.~\ref{fig:data-setup} for an illustration of this protocol.
\begin{itemize}[noitemsep,  topsep=1pt]
    \item {\bf Training}. There are profile images of each instance captured at different views and diverse background images.
    The background images can be used to synthesize training images with free 2D-box annotations, as done by the cut-paste-learn methods~\cite{Lai2014UnsupervisedFL, dwibedi2017cut, georgakis2017synthesizing}.

    \item {\bf Testing}. InsDet algorithms are required to precisely detect all predefined instances from real-world images of cluttered scenes. 
\end{itemize}

{\bf Evaluation metrics}.
The InsDet literature commonly uses average precision (AP) at IoU=0.5~\cite{dwibedi2017cut, ammirato2018target, mercier2021deep};
others use different metrics, 
e.g., AP at IoU=0.75~\cite{hodavn2019photorealistic}, mean AP~\cite{ancha2019combining, georgakis2016multiview}, and F1 score~\cite{bormann2021real}.
As a single metric appears to be insufficient to benchmark methods, we follow the literature of ObjDet that uses multiple metrics altogether~\cite{Lin2014MicrosoftCC}.
\begin{itemize}[noitemsep,  topsep=1pt]
    \item {\bf AP} averages the precision at IoU thresholds from 0.5 to 0.95 with the step size 0.05. It is the \emph{primary metric} in the most well-known COCO Object Detection dataset~\cite{Lin2014MicrosoftCC}.
    \item {\bf AP$_{50}$} and {\bf AP$_{75}$} are the precision averaged over all instances with IoU threshold as 0.5 and 0.75, respectively. In particular, {\bf AP$_{50}$} is the widely used metric in the literature of InsDet.
    \item \note{{\bf AR} (average recall) averages the proposal recall at IoU threshold from 0.5 to 1.0 with the step size 0.05, regardless of the classification accuracy. AR measures the localization performance (excluding classification accuracy) of an InsDet model.}
\end{itemize}
Moreover, we tag \emph{hard} and \emph{easy} scenes in the testing images based on the level of clutter and occlusion, as shown by the right panel of Fig.~\ref{fig:data-setup}.
\note{Following the COCO dataset~\cite{Lin2014MicrosoftCC}, we further tag testing object instances as \emph{small}, \emph{medium}, and \emph{large} according to their bounding box area (cf. details in the supplement).
These tags allow a breakdown analysis to better analyze methods.
}

\subsection{The Dataset}

\begin{wrapfigure}{r}{0.45\textwidth}
\centering
\vspace{-8mm}
\includegraphics[trim={0 0 24cm 0},clip, width=0.86\linewidth]{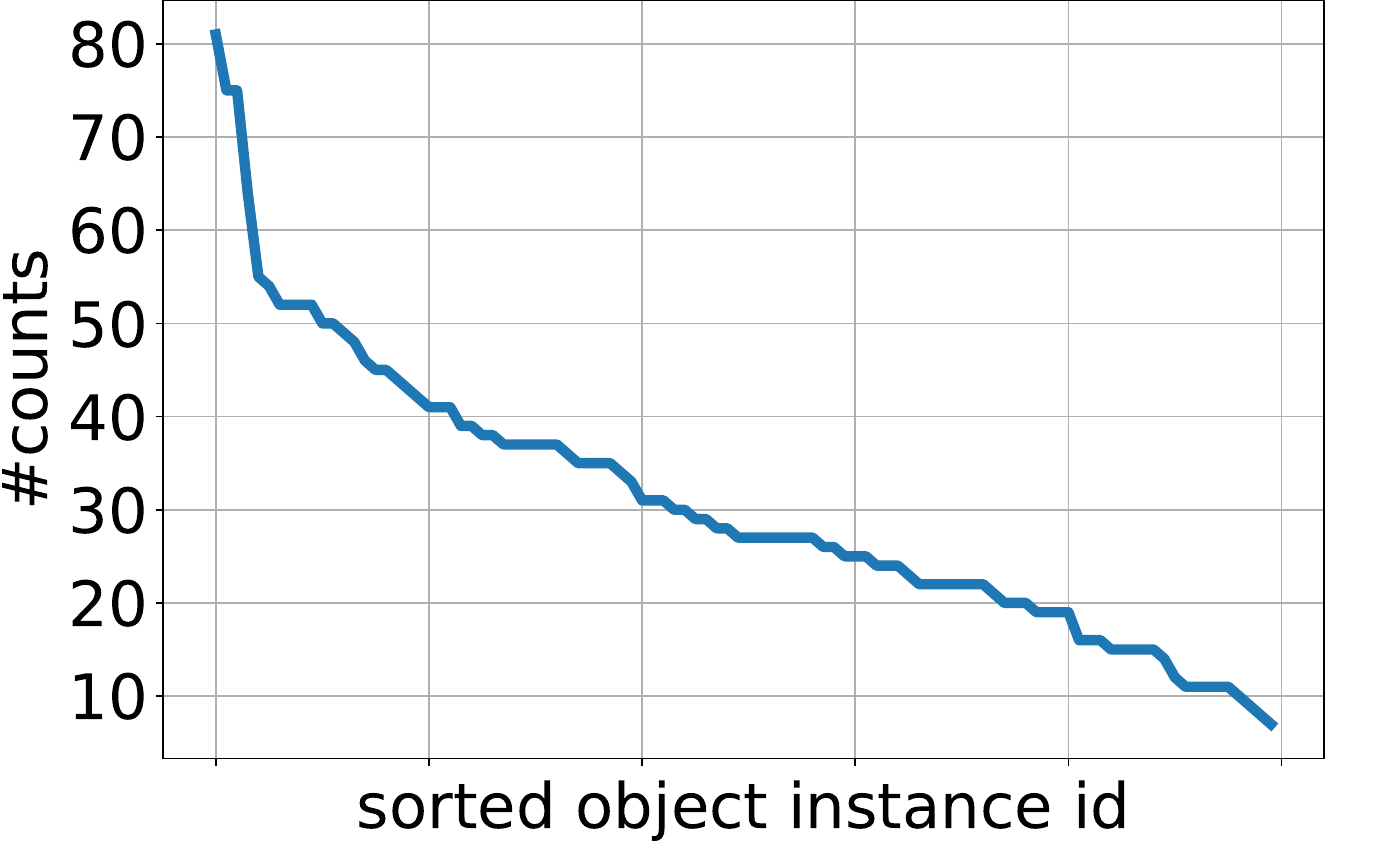}
\vspace{-2mm}
\caption{\small
Imbalanced distribution of instances in test-set.
Yet, instances have the same number of profile images in training and the metrics average over all instances. So, the evaluation is unbiased.
}
\vspace{-3mm}
\label{fig:dist_inst_freq}
\end{wrapfigure}

We introduce a challenging real-world dataset of indoor scenes (\note{motivated by indoor assistive robots}),
including high-resolution photos of 100 distinct object instances, \note{and high-resolution testing images captured from 14 indoor scenes where there are such 100 instances defined for InsDet.}
Table~\ref{tab:datasets} summarizes the statistics compared with existing datasets, showing that our dataset is larger in scale and more challenging than existing InsDet datasets.
Importantly, object instances are located far from the camera in cluttered scenes; this is realistic because robots must detect objects in a distance before approaching them~\cite{active-vision-dataset2017}. Perhaps surprisingly, only a few InsDet datasets exist in the literature. Among them, Grocery~\cite{bormann2021real}, which is the latest and has the most instances like our dataset, is not publicly available.

Our InsDet dataset contains 100 object instances.
When capturing photos for each instance, 
inspired by prior arts~\cite{Singh2014BigBIRDAL, Hinterstoier2012ModelBT, Brachmann2014Learning6O}, we paste a QR code on the tabletop, which enables pose estimation, e.g., using COLMAP~\cite{schonberger2016structure}.
\note{Yet, we note more realistic scenarios can be hand-holding instances for capturing~\cite{Kishida_2021_WACV}, which we think of as future work.}
Each instance photo is of 3072$\times$3072 pixel resolution.
For each instance, we capture 24 photos from multiple views.
The left panel of Fig.~\ref{fig:data-setup} shows some random photos for some instances.
For the testing set, we capture high-resolution images (6144$\times$8192) in cluttered scenes, where some instances are placed in reasonable locations, as shown in the right panel of Fig.~\ref{fig:data-setup}.
We tag these images as \emph{easy} or \emph{hard} based on scene clutter and object occlusion levels. 
When objects are placed sparsely, we tag the testing images as \emph{easy}; otherwise, we tag them as \emph{hard}.
\note{Our InsDet dataset also contains 200 high-res background images of indoor scenes (cf. Fig.~\ref{fig:data-setup}-middle). These indoor scenes are not included in testing images.}
They allow using the cut-paste-learn framework to synthesize training images~\cite{Lai2014UnsupervisedFL, dwibedi2017cut, georgakis2017synthesizing}.
Following this framework, we segment foreground instances using GrabCut~\cite{rother2004grabcut} to paste them on background images.
It is worth noting that the recent vision foundation model SAM~\cite{kirillov2023segment} makes interactive segmentation much more efficient. Yet, this work is made public after we collected our dataset. In Fig.~\ref{fig:dist_inst_freq}, we plot the per-instance frequency in the testing set.

\section{Methodology}

{
\setlength{\tabcolsep}{0.58em} 
\begin{table}[t]
\centering
\small
\caption{\small
{\bf Comparison of our dataset to existing ones}.
Several datasets are used in the InsDet literature although they are designed for different tasks. For example, BigBird and LM are designed to study algorithms of object recognition and object pose estimation, hence they contain instances that are close to the camera. Naively repurposing them for InsDet leads to saturated performance, impoverishing the exploration space of InsDet.
Instead, ours is more challenging as instances are placed far from the camera, simulating realistic scenarios where robots must detect instances at a distance.
Importantly, our dataset contains far more instances than other publicly available InsDet datasets.
}
\begin{tabular}{lcccccccccccccccccccc}
\toprule
 & for what task & publicly available & \#instances & \#scenes &  published year & resolution \\
\midrule 
BigBird~\cite{Singh2014BigBIRDAL}       & {\textcolor{red}{recognition}} &  \cmark  & {\textcolor{darkgreen}{100}} & N/A & 2014 & 1280x1024  \\
RGBD~\cite{Lai2014UnsupervisedFL}     & {\textcolor{red}{scene label.}} & \cmark  & {\textcolor{darkgreen}{300}} & 14 & 2017 & N/A \\
LM~\cite{Hinterstoier2012ModelBT}       & {\textcolor{red}{6D pose est.}}  & \cmark & {\textcolor{red}{15}} & 1  & 2012 &  480x640 \\
LM-O~\cite{Brachmann2014Learning6O}     & {\textcolor{red}{6D pose est.}} & \cmark  & {\textcolor{red}{20}} & 1 & 2017  & 480x640 \\
RU-APC~\cite{rennie2016dataset} & {\textcolor{red}{3D pose est.}} & \cmark  & {\textcolor{red}{14}} & 1 & 2016  & 480x640 \\
\midrule
GMU~\cite{Georgakis2016MultiviewRD}   & {\textcolor{darkgreen}{InsDet}}  & \cmark  & {\textcolor{red}{23}} & 9 & 2016 & 1080x1920 \\ 
AVD~\cite{active-vision-dataset2017}   & {\textcolor{darkgreen}{InsDet}} & \cmark & {\textcolor{red}{33}} & 9 & 2017 & 1080x1920 \\
Grocery~\cite{bormann2021real}      & {\textcolor{darkgreen}{InsDet}}  & \xmark   & {\textcolor{darkgreen}{100}} & 10 & 2021 & unknown \\ 
\midrule
Ours             & {\textcolor{darkgreen}{InsDet}}  & \cmark                     & {\textcolor{darkgreen}{100}} & 14   & 2023 & 6144x8192 \\
\bottomrule
\end{tabular}
\vspace{-1mm}
\label{tab:datasets}
\end{table}
}

\subsection{The Strong Baseline: Cut-Paste-Learn}
\begin{figure*}[t]
\small
\ \hspace{10mm} (a) box \hspace{15mm} (b) Gaussian blurring \hspace{15mm} (c) Motion \hspace{16mm} (d) naive pasting \ \\
\includegraphics[trim=0cm 40cm 0cm 0cm, clip, width=1.0\linewidth]{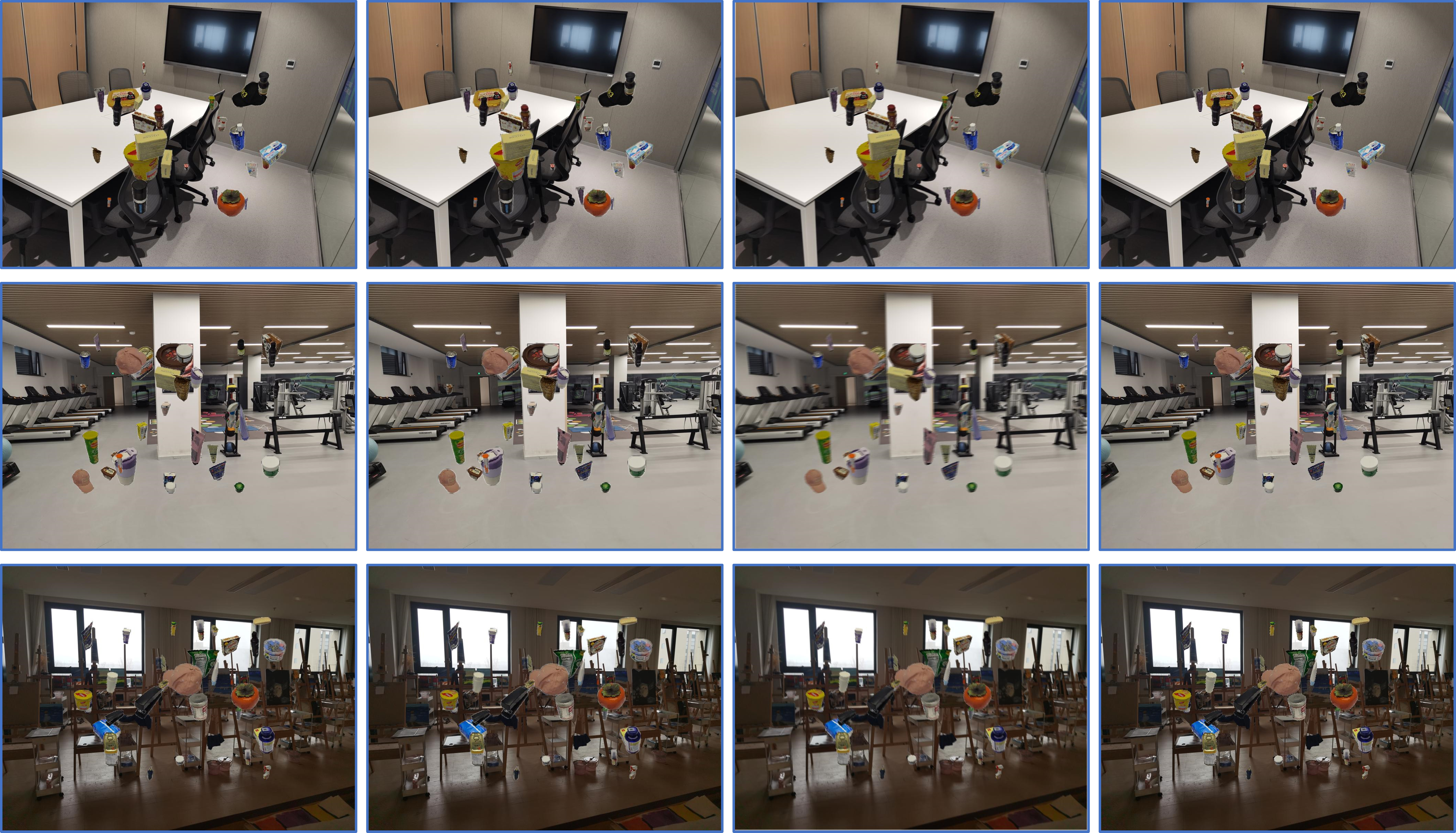}
\vspace{-6mm}
\caption{\small
Synthetic training images for cut-paste-learn methods.
We use different blending methods to paste object instances on the same background. We recommend that interested readers refer to the supplement for an ablation study using different blending methods.
}
\vspace{-3mm}
\label{fig:synthetic-training}
\end{figure*}

{\bf Cut-Paste-Learn} serves as a strong baseline that synthesizes training images with 2D-box annotations~\cite{dwibedi2017cut}. This allows one to train InsDet detectors in the same way as training normal ObjDet detectors, by simply treating the $K$ unique instances as $K$ distinct classes.
It cuts and pastes foreground instances at various aspect ratios and scales on diverse background images, yielding synthetic training images, as shown in Fig.~\ref{fig:synthetic-training}.
Cut-paste-learn is model-agnostic, allowing one to adopt any state-of-the-art detector architecture. 
In this work, we study five popular detectors, 
covering the two-stage detector FasterRCNN~\cite{NIPS2015_fasterRCNN}, and one-stage anchor-based detector RetinaNet~\cite{lin2017focal}, and one-stage anchor-free detectors CenterNet~\cite{zhou2019objects}, and FCOS~\cite{tian2019fcos};
and the transformer-based detector DINO~\cite{zhang2022dino}.
\note{There are multiple factors in the cut-paste-learn framework, such as the number of inserted objects in each background image, their relative size, the number of generated training images and blending methods. We conduct comprehensive ablation studies and report results using the best-tuned choices. We refer interested readers to the supplement for the ablation studies.}

\subsection{The Simple, Non-Learned Method}
We introduce a simple, non-learned InsDet method by exploiting publicly available pretrained models. 
This method consists of three main steps: (1) proposal generation on testing images, (2) matching proposals and profile images,
(3) selecting the best-matched proposals as the detected instances.

{\bf Proposal generation}.
We use the recently released Segment Anything Model (SAM)~\cite{kirillov2023segment} to generate proposals.
For a proposal, we define a minimum bounding square box encapsulating the masked instance, and then crop the region from the high-resolution testing image. 
SAM not only achieves high recall (Table~\ref{tab:RealDB-AR}) on our InsDet dataset but detects objects not belonging to the instances of interest. 
So the next step is to find interested instances from the proposals.

{\bf Feature representation of proposals and profile images}.
Intuitively, among the pool of proposals, we are interested in those that are well-matched to any profile images of any instance.
The well-matched ones are more likely to be predefined instances.
To match proposals and profile images, we use off-the-shelf features to represent them.
In this work, we study two self-supervised learned models as feature extractors, i.e. DINO$_f$~\cite{caron2021emerging}, and DINOv2$_f$~\cite{oquab2023dinov2}.
We feed a square crop (of a proposal) or a profile image to the feature extractor to obtain its feature representation. 
We use cosine similarity over the features as the similarity measure between a proposal and a profile image.

{\bf Proposal matching and selection}.
As each instance has multiple profile images, we need to design the similarity between a proposal and an instance.
For a proposal, we compute the cosine similarities of its feature to all the profile images of an instance and use the maximum as its final similarity to this instance.
We then filter out proposals and instances if they have similarities lower than a threshold, indicating that they are not matched to any instances or proposals.
Finally, we obtain a similarity matrix between all remaining proposals and all remaining instances. 
Over this matrix, we study two matching algorithms to find the best match (hence the final InsDet results), i.e. Rank $\&$ Select, and Stable Matching~\cite{gale1962college, mcvitie1971stable}. 
The former is a greedy algorithm that iteratively selects the best match (highest cosine similarity) between a proposal and an instance and removes the corresponding proposal until no proposal/instance is left.
The latter produces an optimal list of matched proposals and instances, such that there exist no pair of instances and proposals which both prefer each other to their current correspondence under the matching.

\section{Experiments}
\label{sec:experiments}

{\bf Synthesizing training images for cut-paste-learn baselines}. 
Our baseline method trains state-of-the-art ObjDet detectors on  data synthesized using the cut-paste-learn strategy~\cite{dwibedi2017cut}.
For evaluating on our InsDet dataset, we generate 19k training examples and 6k validation examples. For each example, various numbers of foreground objects ranging from 25 to 35 are pasted to a randomly selected background image. The objects are randomly resized with a scale from 0.15 to 0.5. We use four blending options~\cite{dwibedi2017cut}, including Gaussian blurring, motion blurring, box blurring, and naive pasting.
Fig.~\ref{fig:synthetic-training} shows some random synthetic images.
\note{
The above factors have a notable impact on the final performance of trained models, and we have conducted a comprehensive ablation study. We refer interested readers to the supplement for the study.
}

{\bf Implementation details.}
We conduct all the experiments based on open-source implementations, such as Detectron2~\cite{wu2019detectron2} (for FasterRCNN and RetinaNet), CenterNet~\cite{zhou2021probablistic}, FCOS~\cite{tian2019adelaidet} and DINO~\cite{zhang2022dino}. The CNN-based end-to-end detectors are initialized with pretrained weights on COCO~\cite{Lin2014MicrosoftCC}. We fine-tune CNN-based models using SGD and the transformer-based model using AdamW with a learning rate of 1e-3 and a batch size of 16. We fine-tune all the models for 5 epochs (which are enough for training to converge) and evaluate checkpoints after each epoch for model selection. The models are trained on a single Tesla V100 GPU with 32G memory.

If applied, we preprocess object instance profile images and proposals.
Specifically, for a profile image, we remove the background pixels (e.g., pixels of QR code) using foreground segmentation (i.e., GrabCut).
For each proposal, we crop its minimum bounding square box. We also study whether removing background pixels by using SAM's mask output performs better. 
We use DINO$_f$ and DINOv2$_f$ to compute feature representations.

{
\setlength{\tabcolsep}{0.85em} 
\begin{table*}[t]
\centering
\small
\caption{\small
{\bf Benchmarking results on our dataset}.
We summarize three salient conclusions. (1) End-to-end trained detectors perform better with stronger detector architectures, e.g., the transformer DINO (27.99 AP) outperforms FasterRCNN (19.54 AP). 
(2) Interestingly, the non-learned method SAM+DINOv2$_f$ performs the best (41.61 AP), significantly better than end-to-end learned detectors including DINO (27.99 AP).
(3) All methods have much lower AP on {\tt hard} testing images or {\tt small} objects (e.g., SAM+DINOv2$_f$ yields 28.03 AP on {\tt hard} vs. 47.57 AP on {\tt easy}), showing that future work should focus on {\tt hard} situations or {\tt small} instances.
}
\vspace{-2mm}
\begin{tabular}{lcccccccccccccccccccc}
\toprule
 &  \multicolumn{6}{c}{\bf AP} & {\bf AP$_{50}$}  & {\bf AP$_{75}$} \\
\cmidrule(l){2-7}
 & {\tt avg} & {\tt hard} & {\tt easy} & {\tt small} & {\tt medium} & {\tt large} & & \\
\midrule 
FasterRCNN~\cite{NIPS2015_fasterRCNN} 
& \cellcolor{col3}19.54 & 10.26 & 23.75
& 5.03 & 22.20 & 37.97 
& 29.21 & 23.26 
\\
RetinaNet~\cite{lin2017focal}
& \cellcolor{col3}22.22 & 14.92 & 26.49 
& 5.48 & 25.80 & 42.71 
& 31.19 & 24.98 
\\
CenterNet~\cite{zhou2019objects} 
& \cellcolor{col3}21.12 & 11.85 & 25.70 
& 5.90 & 24.15 & 40.38
& 32.72 & 23.60 
\\
FCOS~\cite{tian2019fcos}
& \cellcolor{col3}22.40 & 13.22 & 28.68
& 6.17 & 26.46 & 38.13 
& 32.80 & 25.47 
\\
DINO~\cite{zhang2022dino}
& \cellcolor{col3}27.99 & 17.89 & 32.65 
& 11.51 & 31.60 & 48.35 
& 39.62 & 32.19 
\\
SAM + DINO$_f$
& \cellcolor{col3}36.97 & 22.38 & 43.88 
& 11.93 & 40.85 & 62.67 
& 44.13 & 40.42 
\\
SAM + DINOv2$_f$
& \cellcolor{col3}\textbf{41.61} & \textbf{28.03} & \textbf{47.57} 
& \textbf{14.58} & \textbf{45.83} & \textbf{69.14} 
& \textbf{49.10} & \textbf{45.95} 
\\
\bottomrule
\end{tabular}
\vspace{-1mm}
\label{tab:RealDB}
\end{table*}
}

\subsection{Benchmarking Results}

\begin{figure}[t]
\centering
\begin{minipage}[l]{0.5\textwidth}
\centering
\includegraphics[trim=3cm 0 5cm 0cm, clip, width=1\linewidth]{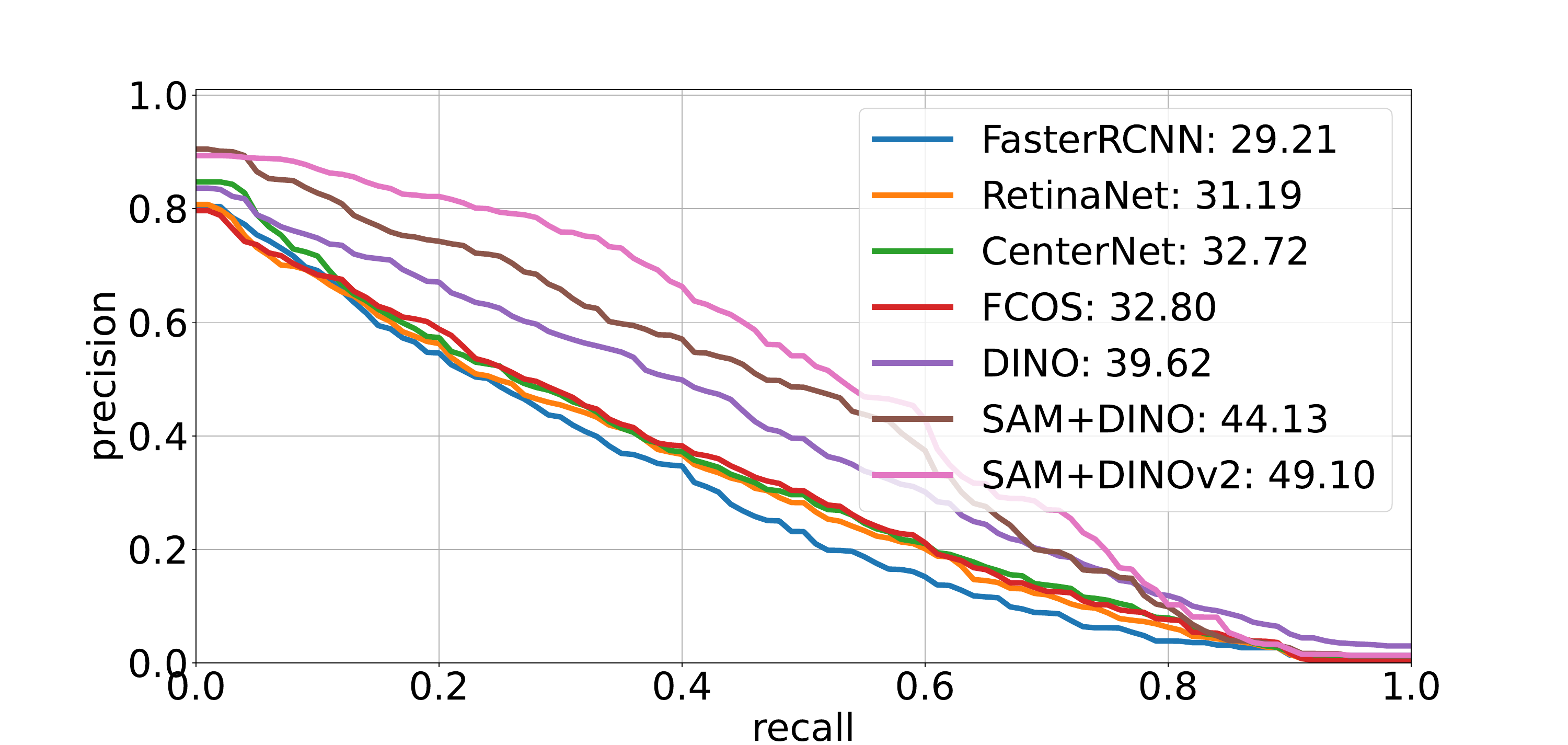}
\end{minipage} \hfill
\begin{minipage}[r]{0.44\textwidth}
\hfill
\caption{\small
Precision-recall curves with IoU=0.5 (AP50 in the legend) on our InsDet dataset.
Stronger detectors perform better, e.g., DINO, a transformer-based detector significantly outperforms FasterRCNN. 
Furthermore, even with a simple non-learned method, leveraging pretrained models, e.g., SAM+DINOv2$_f$, outperforms end-to-end learned methods.
}
\label{fig:PR-curves}
\end{minipage}
\vspace{-3mm}
\end{figure}

{\bf Quantitative results}.
To evaluate the proposed InsDet protocol and dataset, we first train detectors from a COCO-pretrained backbone following the cut-past-learn baseline. 
Table \ref{tab:RealDB} lists detailed comparisons and Fig.~\ref{fig:PR-curves} plots the precision-recall curves for the compared methods.
We can see that detectors with stronger architectures perform better, e.g. DINO (27.99\% AP) vs. FasterRCNN (19.54\% AP).
Second, non-learned methods outperform end-to-end trained models, e.g., SAM+DINOv2$_f$ (41.61\% AP) vs. DINO (27.99\% AP). 
Third, all the methods perform poorly on \emph{hard} and \emph{small} instances, suggesting future work focusing on such cases.

\note{Table~\ref{tab:RealDB-AR} compares methods w.r.t the average recall (AR) metric. ``AR@max10'' means AR within the top-10 ranked detections. 
In computing AR, we rank detections by using the detection confidence scores of the learning-based methods (e.g., FasterRCNN) or similarity scores in the non-learned methods (e.g., SAM+DINO$_f$).
AR$_s$, AR$_m$, and AR$_l$ are breakdowns of AR for small, medium, and large testing object instances.
Results show that (1) the non-learned methods that use SAM generally recall more instances than others, and (2) all methods suffer from small instances. In sum, results show that methods yielding higher recall achieve higher AP metrics (cf. Table~\ref{tab:RealDB}).
}

{
\setlength{\tabcolsep}{0.55em} 
\begin{table*}[t]
\centering
\small
\caption{\small
\note{
{\bf Benchmarking results w.r.t average recall (AR)}.
``AR@max10'' means AR within the top-10 ranked detections. 
In computing AR, we rank detections by using the detection confidence scores of the learning-based methods (e.g., FasterRCNN) or similarity scores in the non-learned methods (e.g., SAM+DINO$_f$).
AR$_s$, AR$_m$, and AR$_l$ are breakdowns of AR for small, medium and large testing object instances.
Results show that (1) the non-learned methods that use SAM generally recall more instances than others, and (2) all methods suffer from small instances. In sum, results show that methods yielding higher recall achieve higher AP metrics (cf. Table~\ref{tab:RealDB}). 
}
}
\vspace{-5mm}
\begin{tabular}{lccccccc}
\toprule
& {\bf AR@max10} & {\bf AR@max100} & {\bf AR$_s$@max100} & {\bf AR$_m$@max100} & {\bf AR$_l$@max100} \\
\midrule
FasterRCNN~\cite{NIPS2015_fasterRCNN} & 26.24 & 39.24
& 14.83 & 44.87 & 60.05
\\
RetinaNet~\cite{lin2017focal} & 26.33 & 49.38
& 22.04 & 56.76 & 69.69
\\
CenterNet~\cite{zhou2019objects} & 23.55 & 44.72
& 17.84 & 52.03 & 64.58
\\
FCOS~\cite{tian2019fcos} & 25.82 & 46.28
& 22.09 & 52.85 & 64.11
\\
DINO~\cite{zhang2022dino} & 29.84 & 54.22
& {\bf 32.00} & 59.43 & 72.92
\\
SAM + DINO$_f$ & 31.25 & 63.05
& 31.65 & 70.01 & {\bf 90.63}
\\
SAM + DINOv2$_f$ & {\bf 40.02} & {\bf 63.06} 
& 31.11 & {\bf 70.40} & 90.36
\\
\bottomrule
\end{tabular}
\vspace{-4mm}
\label{tab:RealDB-AR}
\end{table*}
}

\begin{figure*}[t]
\centering
\includegraphics[width=0.9\linewidth]{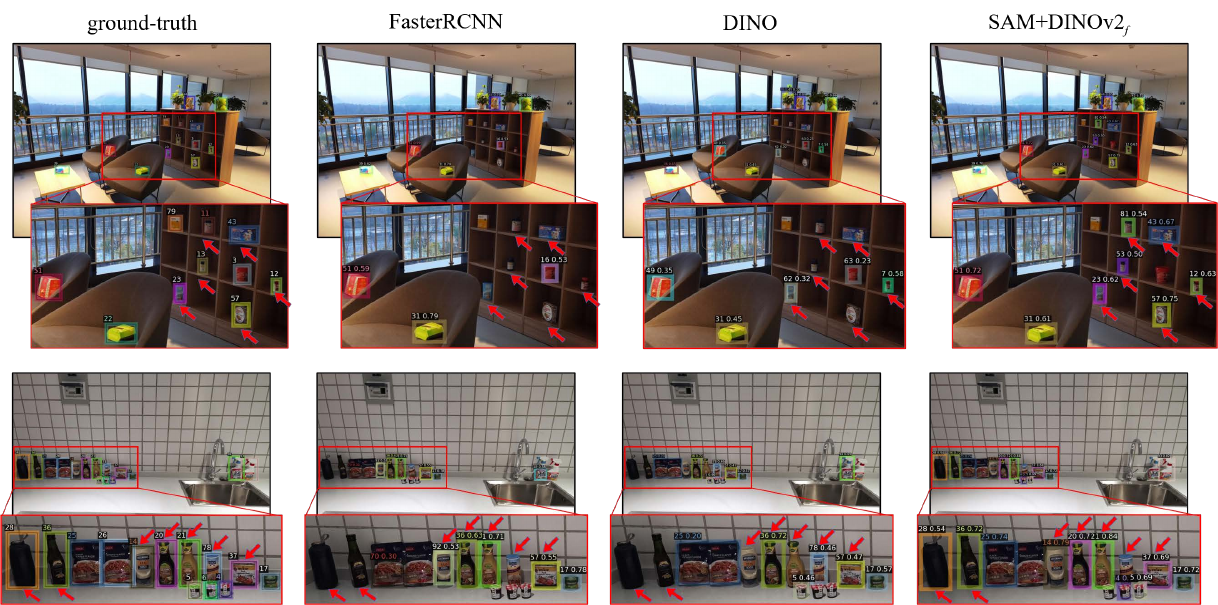}
\vspace{-1mm}
\caption{\small
Visual results of FasterRCNN, DINO, and SAM+DINOv2$_f$ on our InsDet dataset. The top row illustrates the sparse placement of instances (i.e., {\tt easy} scenario), while the bottom contains more cluttered instances (i.e., {\tt hard} scenario). 
We drop predicted instance names for brevity.
SAM helps localize instances with more precise bounding boxes, e.g., as arrows labeled in the upper row. DINOv2$_f$ provides more precise recognition of localized instances, e.g., five instances in the right of the bottom row. 
Compared with DINO, SAM+DINOv2$_f$ is better at locating occluded instances.
}
\vspace{-2mm}
\label{fig:visual-results}
\end{figure*}

{\bf Qualitative results}.
Fig.~\ref{fig:visual-results} visualizes qualitative results on two testing examples from the InsDet dataset.
Stronger detectors, e.g., the non-learned method SAM+DINOv2$_f$, produce fewer false negatives. 
Even so, all detectors still struggle to detect instances with presented barriers such as heavy occlusion, instance size being too small, etc. As shown in Fig.~\ref{fig:PR-curves}, the non-learned method SAM+DINOv2$_f$ outperforms end-to-end learned methods in a wide range of recall thresholds.

{
\setlength{\tabcolsep}{0.6em} 
\begin{table}[t]
\centering
\small
\caption{\small
{\bf Ablation study: whether to remove background in crops for feature computation}.
Based on a proposal given by SAM, we can crop and feed its minimum bounding square to compute DINOv2$_f$ feature, or we can use the mask to remove the background in the square before computing the feature. Clearly, the latter performs remarkably better.
}
\begin{tabular}{lccccccccccc}
\toprule
\multirow{2}*{strategy} &  \multicolumn{3}{c}{\bf AP} & \multicolumn{3}{c}{\bf AP$_{50}$} & \multicolumn{3}{c}{\bf AP$_{75}$} \\
\cmidrule(l){2-4} \cmidrule(l){5-7} \cmidrule(l){8-10}
& {\tt avg} & {\tt hard} & {\tt easy} & {\tt avg} & {\tt hard} & {\tt easy} & {\tt avg} & {\tt hard} & {\tt easy} \\
\midrule 
w/o background removal
& \cellcolor{col3}36.04 & 23.04 & 42.37 
& 43.84 & 29.12 & 51.00 
& 39.59 & 25.74 & 46.13 \\
w/ background removal
& \cellcolor{col3}39.12 & 24.00 & 47.17 
& 46.72 & 30.81 & 54.66 
& 42.86 & 26.40 & 51.58 \\
\bottomrule
\end{tabular}
\vspace{-4mm}
\label{tab:ablation_proposal_generation}
\end{table}
}

{
\setlength{\tabcolsep}{0.75em} 
\begin{table}[t]
\centering
\small
\caption{\small
{\bf Ablation study: whether to generate unique proposal-instance match}. In contrast to Rank\&Select, Stable Matching produces a unique match to proposal/instance for each instance/proposal, yielding better performance than Rank\&Select.
}
\vspace{-0mm}
\begin{tabular}{lcccccccccccccccccccc}
\toprule
\multirow{2}*{strategy} &  \multicolumn{3}{c}{\bf AP} & \multicolumn{3}{c}{\bf AP$_{50}$} & \multicolumn{3}{c}{\bf AP$_{75}$} \\
\cmidrule(l){2-4} \cmidrule(l){5-7} \cmidrule(l){8-10}
& {\tt avg} & {\tt hard} & {\tt easy} & {\tt avg} & {\tt hard} & {\tt easy} & {\tt avg} & {\tt hard} & {\tt easy} \\
\midrule   
Rank $\&$ Select
& \cellcolor{col3}38.62 & 23.95 & 46.31
& 46.04 & 30.77 & 53.64
& 42.37 & 26.39 & 50.61 \\ 
Stable Matching
& \cellcolor{col3}39.12 & 24.00 & 47.17 
& 46.72 & 30.81 & 54.66 
& 42.86 & 26.40 & 51.58 \\
\bottomrule
\end{tabular}
\vspace{-4mm}
\label{tab:ablation_strategies}
\end{table}
}

\subsection{Ablation Study}
\note{
Due to the space limit, we ablate the instance crop and stable matching in the main paper and put more (including ablation studies for the cut-paste-learn methods) in the supplement. 
}

{\bf Proposal feature extraction in the non-learned method.} 
Given a box crop (encapsulating the  proposal) generated by SAM in the non-learned method, we study how to process the crop to improve InsDet performance.
Here, we can either crop and feed its minimum bounding box to compute DINOv2$_f$ features, or we can use the mask to remove the background in the box.
Table~\ref{tab:ablation_proposal_generation} shows the comparison.
Clearly, the latter performs remarkably better in both ``hard'' and ``easy''  scenarios.

{\bf Proposal-instance match in the non-learned method.} After generating proposals by SAM, we need to compare them with instance profile images to get the final detection results. We study the InsDet performance of the two matching algorithms.
Rank $\&$ Select is a greedy algorithm that iteratively finds the best match between any proposals and instances until no instances/proposals are left unmatched;
stable matching produces an optimal list of matched proposals and instances such that there does not exist a pair in which both prefer other proposals/instances to their current correspondence under the matching.
Table \ref{tab:ablation_strategies} compares these two methods, clearly showing that stable matching works better.

\subsection{Discussions}
\label{ssec:discussion}
{\bf Societal Impact}. InsDet is a crucial component in various robotic applications such as elderly-assistive agents. Hence, releasing a unified benchmarking protocol contributes to broader communities.
While our dataset enables InsDet research to move forward, similar to other works, directly applying algorithms brought by our dataset is risky in real-world applications.

{\bf Limitations}.
We note several limitations in our current work.
First, while our work uses normal cameras to collect datasets, we expect to use better and cheaper hardware (e.g., depth camera and IMU) for data collection.
Second, while the cut-paste-learn method we adopt does not consider geometric cues when synthesizing training images, we hope to incorporate such information to generate better and more realistic training images, e.g., pasting instances only on up-surfaces like tables, desks, and floors.
Third, while SAM+DINOv2$_f$ performs the best, this method is time-consuming (see a run-time study in the supplement); real-world applications should consider real-time requirements.

{\bf Future work}.
In view of the above limitations, the future work includes: (1) Exploring high-resolution images for more precise detection on \emph{hard} situations, e.g., one can combine proposals generated from multi-scale and multi-resolution images. (2) Developing faster algorithms, e.g., one can use multi-scale detectors to attend to regions of interest for progressive detection. (3) Bridging end-to-end fast models and powerful yet slow pretrained models, e.g., one can train lightweight adaptors atop pretrained models for better InsDet.

\section{Conclusion}
We explore the problem of Instance Detection (InsDet) by introducing a new dataset consisting of high-resolution images and formulating a realistic unified protocol.
We revisit representative InsDet methods in the cut-paste-learn framework and design a non-learned method by leveraging publicly-available pretrained models.
Extensive experiments show that the non-learned method significantly outperforms end-to-end InsDet models.
Yet, the non-learned method is slow because running large pretrained models takes more time than end-to-end trained models.
Moreover, all methods struggle in hard situations (e.g., in front of heavy occlusions and a high level of clutter in the scene). 
This shows that our dataset serves as a challenging venue for the community to study InsDet.

\bibliographystyle{plainnat}
\bibliography{egbib}

\newpage
\clearpage

\appendix

\begin{center}
{\bf \large Outline}
\end{center}
This document supplements the main paper with more experimental results,
more visualizations, 
further details of the InsDet dataset, and open-source code. 
Below is the outline of this document.

\begin{itemize} [noitemsep, topsep=-1pt, leftmargin=*]    

\item
    {\bf Section~\ref{sec:jupyter-notebook}}.
    We provide demo code for the non-learned method using Jupyter Notebook.
    
\item 
    {\bf Section~\ref{sec:vis}}.
    We visualize object instance profile images captured in multiple views, instance proposals produced by SAM, and input crops to DINO$_f$ / DINOv2$_f$.

\item 
    {\bf Section~\ref{sec:dataset}}. We display how to tag object instances as ``small'', ``medium'' and ``large'', and supplement the average recall of proposals under both ``easy'' and ``hard'' scenes.  
\item 
    {\bf Section~\ref{sec:ablation}}. We conduct extensive ablation studies on the non-learned InsDet method and the traditional cut-paste-learn method. 

\item 
    {\bf Section~\ref{sec:comparison}}
    We demonstrate runtime comparison between end-to-end learned detectors and non-learned detectors, and discuss their trade-off.

\item 
    {\bf Section~\ref{sec:datasheets for datasets}} includes dataset documentation and intended uses.
    
\item {\bf Appendix} contains further details of dataset collection.
    
\end{itemize}

\section{Open-Source Code}
\label{sec:jupyter-notebook}

We release open-source code in the form of Jupyter Notebook plus Python files.

{\bf Why Jupyter Notebook?}
We prefer to release the code using Jupyter Notebook (\url{https://jupyter.org}) because it allows for interactive demonstration for education purposes.
In case the reader would like to run Python script, using the following command can convert a Jupyter Notebook file {\tt xxx.ipynb} into a Python script file {\tt xxx.py}:
{\small 
\begin{verbatim}jupyter nbconvert --to script xxx.ipynb\end{verbatim}
}

{\bf Requirement}.
Running our code requires some common packages.
We installed Python and most packages through Anaconda. A few other packages might not be installed automatically, such as Pandas, torchvision, and PyTorch, which are required to run our code. Below are the versions of Python and PyTorch used in our work. 
\begin{itemize}
\item Python version: 3.9.16 [GCC 7.5.0]
\item PyTorch version: 2.0.0
\end{itemize}
We suggest assigning $>$30GB space to run all the files.

{\bf License}.
We release open-source code under the MIT License to foster future research in this field.

{\bf Demo.}
The Jupyter notebook files below demonstrate our non-learned method using SAM and DINOv2$_f$. The masked instances generated by SAM are encapsulated by a minimum bounding square box and then cropped from the high-resolution testing image. We feed these proposals into DINOv2$_f$ for feature representation, just like how we represent  profile images. We run Rank $\&$ Select and Stable Matching to determine ``well-matched'' proposals and instances.

\begin{itemize}
\item
\begin{verbatim}demo_get_proposals.ipynb\end{verbatim}
Running this file crops a masked instance from the high-resolution testing image with/without background, and visualizes the cropped proposal regions on the testing image.

\item
\begin{verbatim}demo_eval_instance_detection_Stable_Matching.ipynb\end{verbatim}
Running this file extracts features of proposals and object instance profile images by DINOv2$_f$, and implements Stable Matching to return matched proposals and instances.

\item
\begin{verbatim}demo_eval_instance_detection_Rank_Select.ipynb\end{verbatim}
Running this file extracts features of proposals and object instance profile images by DINOv2$_f$, and implements Rank $\&$ Select to return matched proposals and instances.
\end{itemize}

\section{More Visualizations}
\label{sec:vis}

\begin{figure*}[t]
\small
\includegraphics[trim={0 0 0 131mm},clip, width=1\linewidth]{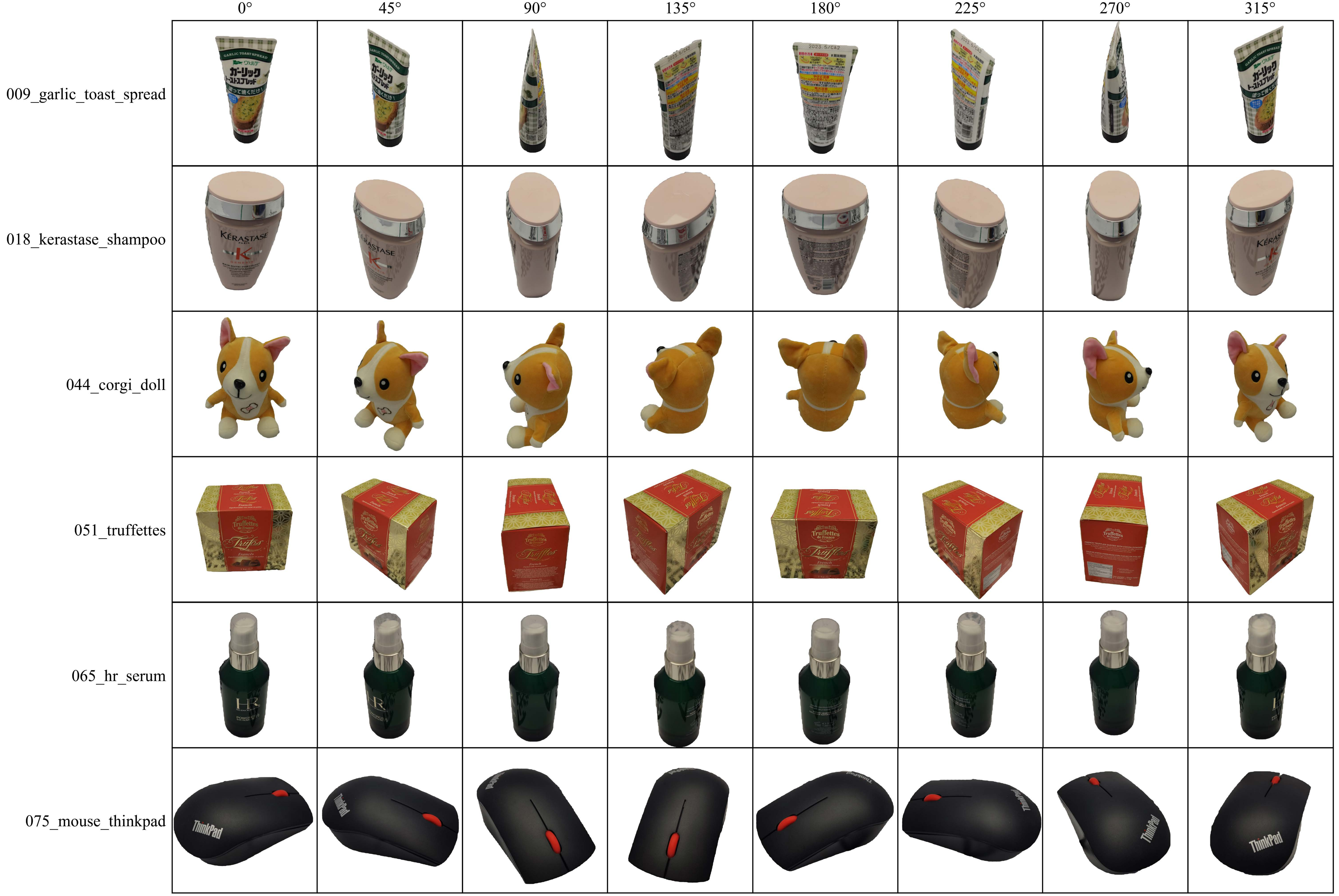}
\vspace{-4mm}
\caption{\small
Examples of object instance profile images.
We demonstrate profile images of five random object instances (after background removal using GrabCut).
}
\label{fig:vis-instances}
\end{figure*}

{\bf Instance Profile Images}.
Fig.~\ref{fig:vis-instances} presents more visualizations of object instance profile images in our InsDet dataset. The InsDet dataset contains 100 object instances, including sauces, snack foods, office stationery, cosmetics, toiletries, dolls, etc. 
Each instance is captured at 24 rotation positions (every 15$^\circ$ in azimuth) with 45$^\circ$ elevation view.
Profile images are captured at 3072$\times$3072 pixel resolution (some are  3456$\times$3456).
We use the GrabCut~\cite{rother2004grabcut} toolbox to derive foreground masks of instances in profile images.
This removes background pixels (such as QR code regions) in the profile images. 
In practice, we center-crop foreground instances from profile images and downsize the center-crops to 1024$\times$1024.
As shown in Fig.~\ref{fig:vis-instances}, we visualize object instances of various shapes and sizes in multiple rotation views.

\begin{figure*}[t]
\centering
\includegraphics[width=0.45\linewidth]{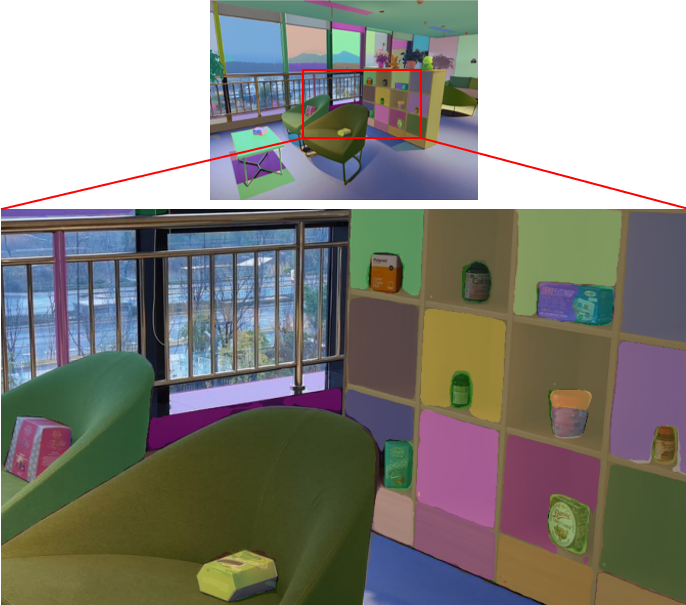} \hfill
\includegraphics[width=0.45\linewidth]{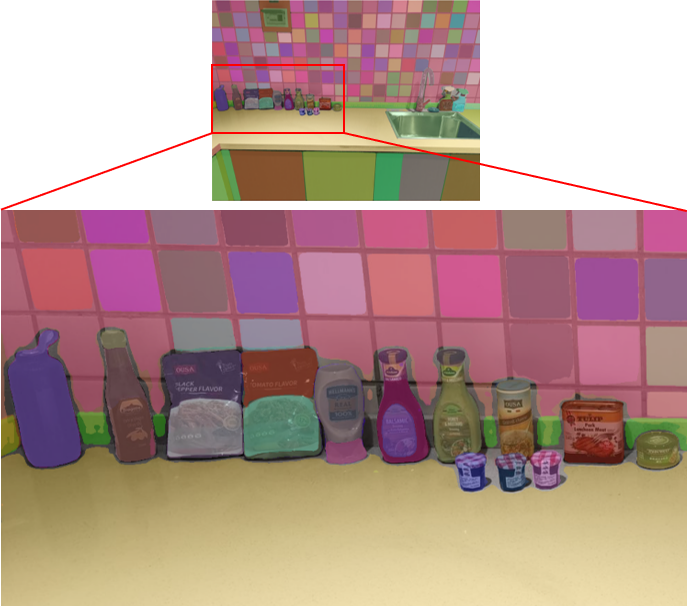}
\vspace{-2mm}
\caption{\small
Instance proposals produced by SAM overlaid on the testing images.
Note that at each location seed, SAM produces three proposals which might be object parts or the whole object. We use all of them as proposal detections and let the follow-up step of Stable Matching find the best-matched proposals to instances as the final InsDet results.
}
\label{fig:SAM-proposals}
\end{figure*}

{\bf Instance Proposals Produced by SAM}.
Fig.~\ref{fig:SAM-proposals} shows the segmentation masks produced by SAM on two testing scene images (demonstrated on page 8 of the paper). We downsize the high-resolution testing images to low-resolution (e.g., 768$\times$1024 or 1536$\times$2048) for more efficient and effective segmentation. We observe that SAM can produce high-quality segments, although it also over-segments.
Note that at each location seed, SAM produces three proposals which might be object parts or the whole object. We use all of them as proposed detections and let the follow-up step of Stable Matching find the best-matched proposals to instances as the final InsDet results.

\begin{figure*}[t]
\centering
\small
\ \hspace{15mm} (a) w/ background \hspace{48mm} (b) w/o background \hspace{8mm} \ \\
\includegraphics[width=0.47\linewidth]{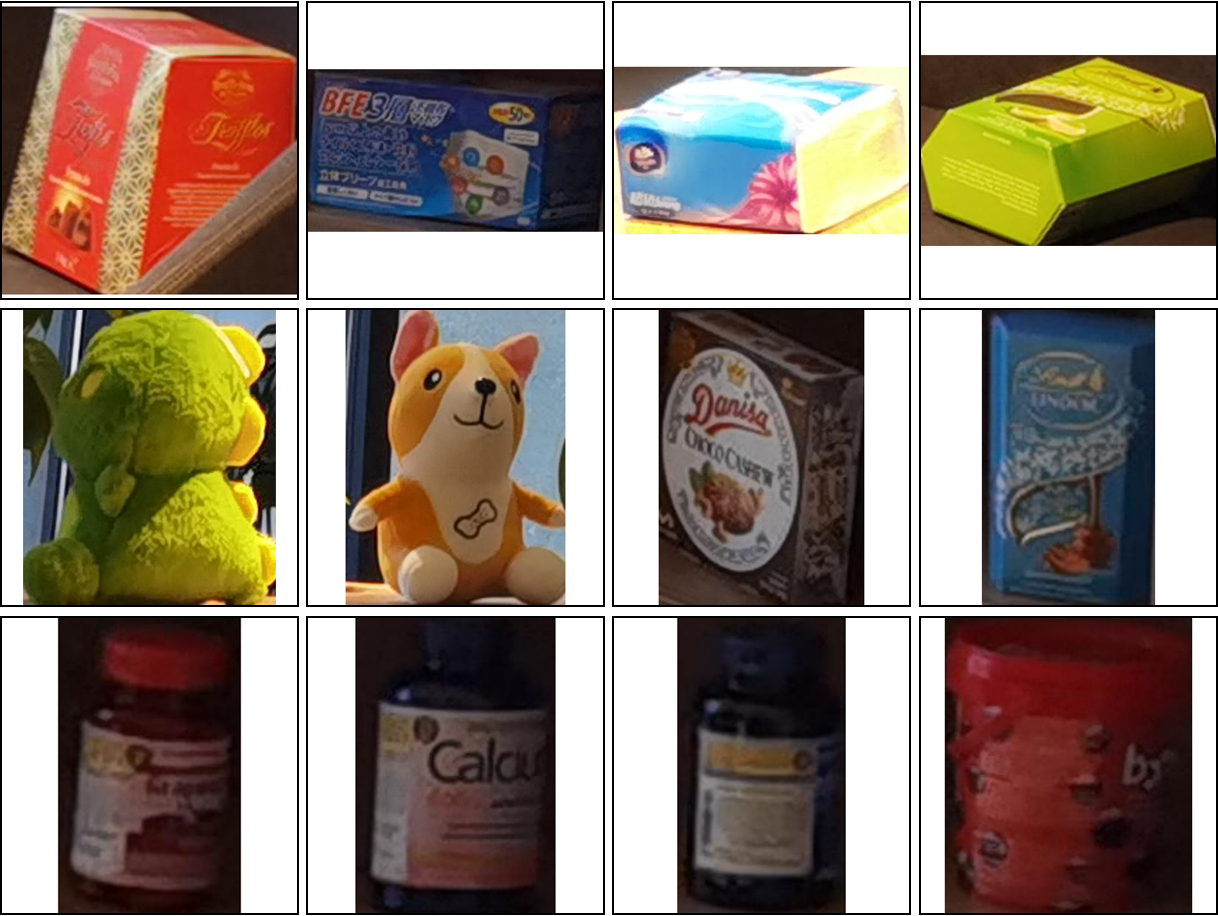}\hfill
\includegraphics[width=0.47\linewidth]{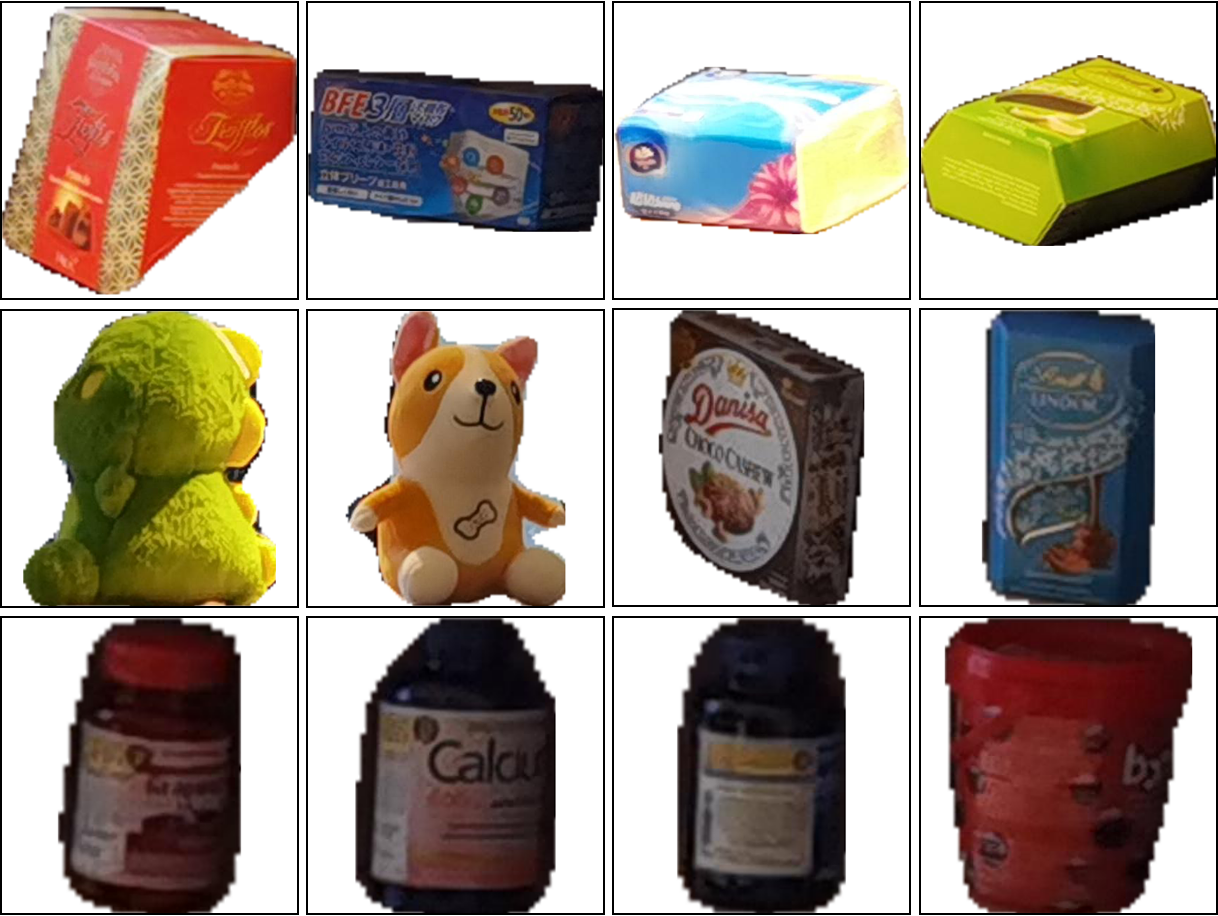}
\vspace{-2mm}
\caption{\small
Examples of input proposals to DINOv2$_f$. Based on a proposal given by SAM, we crop the instance using the minimum bounding square ((a) w/ background) or using the segmentation mask ((b) w/o background) before computing the features.
}
\vspace{-1mm}
\label{fig:DINOv2_input}
\end{figure*}

{\bf Input Proposals to DINOv2$_f$}.
Fig.~\ref{fig:DINOv2_input} presents two strategies of feeding proposals to DINOv2$_f$ for feature computation. One is to crop an instance proposal with its minimum bounding square which keeps the background from the testing image. Another is to use such with background removal (via the mask generated by SAM).

\begin{figure}[t]
\centering
\includegraphics[width=0.48\textwidth]{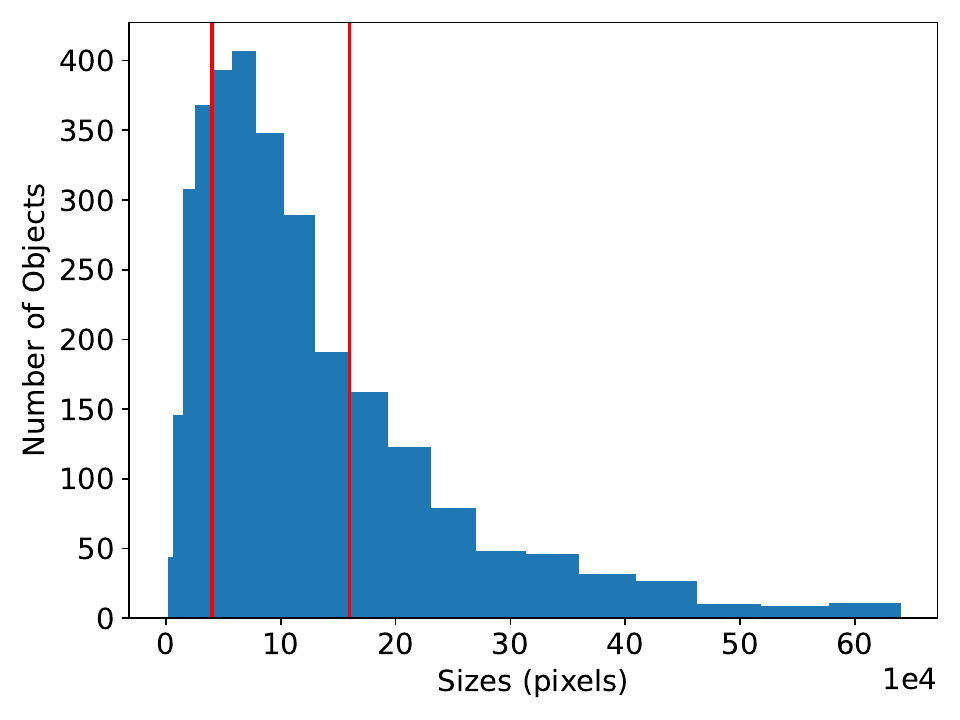}\\
\vspace{-3mm}
\caption{
Distribution of objects w.r.t their bounding box area in testing images. We split them into {\tt small}, {\tt medium}, and {\tt large} subgroups to allow breakdown analysis.
}
\label{fig:dataset-barplot}
\end{figure}

\begin{table}[t]
\centering
\caption{Following the spirit of COCO dataset, we tag objects with different sizes by {\tt small}, {\tt medium}, and {\tt large}, respectively. }
\begin{tabular}{l c }
\hline
size & bounding box area\\
\hline 
small & $<$ 200$^2$ \\  
\hline
medium & 200$^2$ - 400$^2$\\  
\hline
large & $>$ 400$^2$ \\  
\hline
\end{tabular}
\label{tab:dataset-ObjSize}
\end{table}

{
\setlength{\tabcolsep}{1.5em} 
\begin{table*}[!ht]
\centering
\small
\caption{\small
{\bf Benchmarking results w.r.t average  recall (AR)}.
We added a breakdown analysis of testing images on {\tt hard} and {\tt easy} scenes.
Results show that (1) the non-learned methods that use SAM generally recall more instances than others, and (2) all methods suffer from hard scenes.
}
\vspace{-2mm}
\begin{tabular}{lccccccc}
\toprule
& \multicolumn{3}{c}{\bf AR@max10} & \multicolumn{3}{c}{\bf AR@max100} \\
\cmidrule(l){2-4} \cmidrule(l){5-7}
& {\tt avg} & {\tt hard} & {\tt easy} & {\tt avg} & {\tt hard} & {\tt easy} \\
\midrule
FasterRCNN~\cite{NIPS2015_fasterRCNN} & 26.24 & 12.92 & 32.33 & 39.24 & 16.91 & 49.43
\\
RetinaNet~\cite{lin2017focal} & 26.33 & 15.38 & 31.33 & 49.38 & 29.00 & 58.69
\\
CenterNet~\cite{zhou2019objects} & 23.55 & 11.87 & 28.87 & 44.72 & 24.88 & 53.76
\\
FCOS~\cite{tian2019fcos} & 25.82 & 12.81 & 31.74 & 46.28 & 26.55 & 55.27
\\
DINO~\cite{zhang2022dino} & 29.84 & 16.63 & 35.84 & 54.22 & 36.46 & 62.30
\\
SAM + DINO$_f$ & 31.25 & 16.96 & 37.73 & 63.05 & 42.46 & {\bf 72.41}
\\
SAM + DINOv2$_f$ & {\bf 40.02} & {\bf 27.64} & {\bf 45.36} & {\bf 63.06} & {\bf 43.47} & 71.96
\\
\bottomrule
\end{tabular}
\vspace{-4mm}
\label{tab:RealDB-AR-hard-easy}
\end{table*}
}

\section{Dataset Statistics}
\label{sec:dataset}

In addition to tag testing images by \emph{hard} and \emph{easy} according to the level of scene clutter and object occlusion, we further tag testing instances based on their size. Specially, following the spirit of COCO dataset, we tag instances with \emph{small}, \emph{medium}, and \emph{large}, as in Table~\ref{tab:dataset-ObjSize}. 
To determine their size tags, we plot the distribution of their sizes in Fig.~\ref{fig:dataset-barplot}, showing an intuitive way to tag them.

As a supplement to Table 3 in the main paper showing the average recall (AR) of \emph{small}, \emph{medium} and \emph{large}, Table~\ref{tab:RealDB-AR-hard-easy} further studies AR in \emph{hard} and \emph{easy} scenes. We can observe that: (1) the non-learned methods that use SAM recall more proposals than other competitors in both \emph{easy} and \emph{hard} scenes; (2) all methods basically suffer from \emph{hard} scenes.

\section{Ablation Study}
\label{sec:ablation}

We conduct more ablation studies on the non-learned InsDet method and the traditional cut-paste-learn method. In the cut-paste-learn group of methods, there are four factors influencing the final detection results, i.e., the number of objects inserted per image, the scales of inserted object instances, the blending methods when pasting instances on background images, and the total amount of synthesized training images. We ablate these factors by using the FasterRCNN architecture.

{\bf Impact of different image/proposal resolutions.}
We study the InsDet performance when using images of different sizes for SAM, and using different resolutions of crops fed into DINOv2$_f$. 
For example, we can use SAM on images of size 3072$\times$4096 to generate proposals, we can resize crops of proposals to 224$\times$224 to feed into DINOv2$_f$ if they are larger than 224$\times$224 (otherwise, keep them unchanged).
Table~\ref{tab:ablation_proposal_crops} lists detailed comparisons.
We have two observations.
(1) The InsDet performance generally increases with the image resolution but starts to drop when the input image is too large, i.e., 6144$\times$8192. This is because SAM tends to produce object parts as individual instances, resulting in more false positives. 
(2) When using larger proposals for DINOv2$_f$, InsDet performance gets better, e.g., 41.61\% (448$\times$448) vs. 39.12\% (224$\times$224).

{
\setlength{\tabcolsep}{0.48em} 
\begin{table}[!htp]
\centering
\small
\caption{\small
{\bf Ablation study: which image/proposal size to use for SAM and DINOv2$_f$. } We notice that InsDet performance generally increases with the input image resolution, but starts to drop when the image is too large. When using larger proposals for DINOv2$_f$, InsDet performance also gets better. 
}
\vspace{-0mm}
\begin{tabular}{ccccccccccccccccccccc}
\toprule
\multirow{2}*{image resolution} & \multirow{2}*{input size} & \multicolumn{3}{c}{\bf AP} & \multicolumn{3}{c}{\bf AP$_{50}$} & \multicolumn{3}{c}{\bf AP$_{75}$} \\
\cmidrule(l){3-5} \cmidrule(l){6-8} \cmidrule(l){9-11}
for SAM & for DINOv2$_f$ & {\tt avg} & {\tt hard} & {\tt easy} & {\tt avg} & {\tt hard} & {\tt easy} & {\tt avg} & {\tt hard} & {\tt easy} \\
\midrule 
768$\times$1024 & \multirow{4}*{224$\times$224}
& \cellcolor{col3}36.46 & 21.67 & 43.88  
& 46.22 & 29.52 & 54.11 
& 41.53 & 24.95 & 49.99 \\
1536$\times$2048 & 
& \cellcolor{col3}39.12 & 24.00 & 47.17 
& 46.72 & 30.81 & 54.66 
& 42.86 & 26.40 & 51.58 \\
3072$\times$4096 & 
& \cellcolor{col3}39.17 & 24.08 & 46.60 
& 45.71 & 30.34 & 53.07 
& 41.90 & 26.12 & 49.71 \\
6144$\times$8192 & 
& \cellcolor{col3}38.74 & 23.39 & 46.29  
& 45.24 & 29.24 & 52.78 
& 40.81 & 25.14 & 48.65 \\
\midrule
\multirow{3}*{1536$\times$2048}& 112$\times$112
& \cellcolor{col3}26.46 & 16.52 & 31.32 
& 30.83 & 20.94 & 36.26 
& 28.89 & 18.81 & 33.73 \\
& 224$\times$224
& \cellcolor{col3}39.12 & 24.00 & 47.17 
& 46.72 & 30.81 & 54.66 
& 42.86 & 26.40 & 51.58 \\
& 448$\times$448
& \cellcolor{col3}41.61 & 28.03 & 47.57  
& 49.10 & 36.64 & 54.84 
& 45.95 & 31.41 & 52.03 \\
\bottomrule
\end{tabular}
\vspace{-1mm}
\label{tab:ablation_proposal_crops}
\end{table}
}

{
\setlength{\tabcolsep}{0.75em} 
\begin{table*}[t]
\centering
\small
\caption{\small
{\bf Ablation study: number of objects inserted in each background image}. Basically, inserting more objects helps train InsDet detectors and achieves better performance. Yet, inserting more is not necessarily increasing much further.
}
\vspace{-2mm}
\begin{tabular}{lcccccc}
\toprule
 $\#$ of objects &  AP &AP$_{50}$  & AP$_{75}$ & AP$_s$ & AP$_m$ & AP$_l$ \\
\midrule
$[5,15]$ 
& 17.57 & 25.98 & 20.49 & 3.55 & 20.31 & 33.50 \\
$[15,25]$
& 18.20 & 27.76 & 21.09 & 4.45 & 20.66 & 35.51 \\
$[25,35]$
& 19.39 & 29.14 & {\bf 23.09} & 5.03 & 22.04 & 37.73 \\
$[35,45]$
& {\bf 19.60} & {\bf 30.30} & 22.82 & {\bf 5.44} & {\bf 22.32} & {\bf 39.17} \\
\bottomrule
\end{tabular}
\label{tab:ablation-objnum}
\end{table*}
}

{
\setlength{\tabcolsep}{0.75em} 
\begin{table*}[!ht]
\centering
\small
\caption{\small
{\bf Ablation study: scales of inserted object instances in synthesizing training images}. The scale significantly influences the final detection performance. Inserting objects that are too small (e.g. [0.1, 0.15]) or too large (e.g. [0.5, 1.0]) will not train detectors well. We think this is because the testing images contain more ``medium'' object instances.
}
\vspace{-2mm}
\begin{tabular}{lcccccc}
\toprule
 scale of objects &  AP &AP$_{50}$  & AP$_{75}$ & AP$_s$ & AP$_m$ & AP$_l$ \\
\midrule
$[0.1,0.15]$ 
& 4.72 & 9.63 & 4.16 & 5.48 & 8.93 & 0.72 \\
$[0.15,0.3]$
& 16.66 & 26.82 & 18.55 & {\bf 16.01} & {\bf 27.74} & 9.88 \\
$[0.15,0.5]$
& {\bf 19.39} & {\bf 29.14} & {\bf 23.09} & 5.03 & 22.04 & 37.73 \\
$[0.5,0.8]$
& 5.43 & 8.16 & 6.08 & 1.79 & 18.72 & {\bf 70.20} \\
$[0.5,1.0]$
& 5.74 & 9.15 & 6.60 & 0.00 & 3.00 & 19.58 \\
\bottomrule
\end{tabular}
\label{tab:ablation-scale}
\end{table*}
}

{
\setlength{\tabcolsep}{0.75em} 
\begin{table*}[!ht]
\centering
\small
\caption{\small
{\bf Ablation study: blending methods between objects and the background images.} 
We note that: (1) Naive pasting gives the worst performance, since directly pasting objects on background images creates boundary artifacts. (2) Although the other three blending modes do not yield visually perfect results, they could still improve the detection performance. (3) When using all four blending modes when mixing the same background image with the same object displacement, the InsDet performance could be further improved. This is because training on multiple images makes the algorithm less sensitive to these blending factors. 
}
\vspace{-2mm}
\begin{tabular}{lcccccc}
\toprule
 blending strategy &  AP &AP$_{50}$  & AP$_{75}$ & AP$_s$ & AP$_m$ & AP$_l$ \\
\midrule
Gaussian 
& 17.83 & 27.13 & 21.02 & 4.74 & 20.49 & 36.09 \\
motion
& 17.92 & 27.57 & 20.85 & 4.69 & 20.76 & 34.78 \\
box blurring
& 17.71 & 27.47 & 20.95 & 4.25 & 20.30 & 34.56 \\
naive pasting
& 16.53 & 24.77 & 20.17 & 4.25 & 19.07 & 34.86 \\
all
& {\bf 19.39} & {\bf 29.14} & {\bf 23.09} & {\bf 5.03} & {\bf 22.04} & {\bf 37.73} \\
\bottomrule
\end{tabular}
\vspace{-1mm}
\label{tab:ablation-blending}
\end{table*}
}
\note{
{\bf Number of objects inserted in each background image.} 
We study InsDet performance with different numbers of objects inserted in each background image in Table~\ref{tab:ablation-objnum}. We can see that inserting more objects helps train InsDet detectors and achieves better performance. Concretely, FasterRCNN yields 19.54\% AP when trained on synthesized training images each of which has 25-35 object instances, better than 17.57\% AP when trained on those that have 5-15 object instances per image. But inserting more is not necessarily increasing much further.}

\note{
{\bf Scales of inserted object instances in synthesizing training images.} 
We study the impact of the scales of inserted object instances when synthesizing training images in Table~\ref{tab:ablation-scale}. The number in square brackets denotes the range of downsampling factors for instance profile images. For example, $[0.1, 0.15]$ denotes that the original instance profile images (256x256 resolution) are randomly scaled by 0.1-0.15 before being pasted on background images. We can see that the scale significantly influences the final detection performance. For example, inserting objects that are too small (e.g. [0.1, 0.15]) or too large (e.g. [0.5, 1.0]) will not train detectors well. We conjecture this is because the testing images contain more ``medium'' object instances.}

\note{
{\bf Blending methods between objects and background.} 
We study four commonly-used blending methods when pasting objects into the background images in Table~\ref{tab:ablation-blending}, i.e. Gaussian blurring, motion blurring, box blurring, and naive pasting.  We note that: (1) naive pasting yields the worst performance, since this creates boundary artifacts; (2) the other three blending methods work better than naive pasting but do not show significant performance difference; (3) using all the four blending methods together leads to the best performance, significantly better than using any one of them alone.}

\note{
{\bf Total amount of synthesized training images. } 
We study InsDet performance by training on different amounts of synthesized images in Table~\ref{tab:ablation-training_images}. Perhaps surprisingly, using 5k synthesized training images is better than training on more images! We conjecture the reasons are that (1) more synthesized images do not bring new signals to help training, (2) domain gaps between synthesized images and real testing images are difficult to close by simply using more such synthetic data, otherwise, training will overfit to them and hence hurt the final InsDet performance. }

{
\setlength{\tabcolsep}{0.75em} 
\begin{table*}[t]
\centering
\small
\caption{\small
{\bf Ablation study: different amounts of synthesized training images}.
Perhaps surprisingly, using 5k synthesized training images is better than training on more images! We conjecture the reasons are that (1) more synthesized images do not bring new signals to help training, (2) domain gaps between synthesized images and real testing images are difficult to close by simply using more such synthetic data, otherwise, training will overfit to them and hence hurt the final InsDet performance.
}
\vspace{-2mm}
\begin{tabular}{lcccccc}
\toprule
 $\#$ of training images &  AP & AP$_{50}$  & AP$_{75}$ & AP$_s$ & AP$_m$ & AP$_l$ \\
\midrule
5k 
& {\bf 19.93} & {\bf 30.60} & {\bf 23.21} & {\bf 5.62} & 22.09 & {\bf 38.92} \\
10k
& 19.08 & 29.50 & 21.91 & 4.67 & 21.65 & 37.21 \\
20k
& 19.39 & 29.14 & 23.09 & 5.03 & 22.04 & 37.73 \\
25k
& 19.19 & 29.13 & 22.33 & 4.46 & {\bf 22.21} & 36.92 \\
30k
& 18.42 & 28.11 & 21.42 & 4.38 & 21.19 & 36.70 \\
\bottomrule
\end{tabular}
\vspace{-1mm}
\label{tab:ablation-training_images}
\end{table*}
}

\begin{table}[t]
\centering
\begin{minipage}[l]{0.38\textwidth}
\centering
\vspace{4mm}
\small
{
\setlength{\tabcolsep}{0.8em}
\begin{tabular}{l c c}
\hline
method &  time (sec) & AP (\%)\\
\hline 
FasterRCNN~\cite{NIPS2015_fasterRCNN} & 0.00399 & 19.54\\  
RetinaNet~\cite{lin2017focal} & 0.00412  & 22.22 \\  
CenterNet~\cite{zhou2019objects} & 0.00376  & 21.12 \\  
FCOS~\cite{tian2019fcos} & 0.00271 & 22.40 \\  
DINO~\cite{zhang2022dino} & 1.90625 & 27.99 \\   
SAM + DINO$_f$ & 15.10 \ \ \ \ \ & 36.97 \\
SAM + DINOv2$_f$ & 14.70 \ \ \ \ \  & 41.61 \\  
\hline
\end{tabular}
}
\end{minipage} 
\hfill
\begin{minipage}[r]{0.54\textwidth}
\vspace{-1mm}
\caption{\small
We compare the inference runtime (second/image) of different methods, along with their InsDet performance in AP (\%).
Clearly, there is a trade-off between runtime and detection precision.
For example, among the methods studied in our work, SAM+DINOv2$_f$ achieves the highest AP (41.61) but is four orders of magnitude slower than FasterRCNN (19.54 AP).
Developing faster and better InsDet methods is apparently future work.
}
\label{tab:comparison}
\end{minipage}
\end{table}

\section{Runtime Comparison}
\label{sec:comparison}

Table~\ref{tab:comparison} compares the runtime of different methods, along with their InsDet performance.
There is a trade-off between runtime and detection precision.
For example, among the methods studied in our work, SAM+DINOv2$_f$ achieves the highest AP (41.61) but is four orders of magnitude slower than FasterRCNN (19.54 AP).
Developing faster and better InsDet methods is future work.

\section{Datasheet for our Dataset}
\label{sec:datasheets for datasets}

We follow the datasheet proposed in~\cite{gebru2021datasheets} for documenting our InsDet dataset.
\begin{enumerate}
    \item Motivation
    \begin{enumerate}
    \item For what purpose was the dataset created? \\
    This dataset was created to study the problem of Instance Detection, i.e., detecting individual object instances from every single-image of cluttered scenes. 
    \item Who created the dataset and on behalf of which entity? \\
    This dataset was mainly created by Qianqian Shen. Other authors help with logistics.
    \item Who funded the creation of the dataset? \\
    \answerNA{}
    \item Any other Comments? \\
    \answerNo{}
    \end{enumerate}
    \item Composition
    \begin{enumerate}
        \item What do the instances that comprise the dataset represent? \\
        RGB images captured by a camera, and annotation files.
        \item How many instances are there in total? \\
        There are 2,760 instances including 24*100 profile images (100 objects with each of which having 24 profile images), 200 background images, and 160 testing images.
        \item Does the dataset contain all possible instances or is it a sample (not necessarily random) of instances from a larger set? \\
        It contains all possible instances.
        \item What data does each instance consist of? \\
        See 2.(a).
        \item Is there a label or target associated with each instance? \\
        See 2.(a)
        \item Is any information missing from individual instances? \\
        \answerNo{}
        \item Are relationships between individual instances made explicit? \\
        \answerYes{Images and annotations files are associated.}
        \item Are there recommended data splits? \\
        \answerYes{We provide training and testing split for the dataset.}
        \item Are there any errors, sources of noise, or redundancies in the dataset? \\
        \answerNo{We tried our best to manually annotate bounding boxes of object instances on the testing image, and we did not see visible noise or errors. In profile images of instances, there might be noise in camera pose estimation due to the QR code pasted on the table.}
        \item Is the dataset self-contained, or does it link to or otherwise rely on external resources (e.g., websites, tweets, other datasets)? \\
        \answerYes{}
        \item Does the dataset contain data that might be considered confidential (e.g., data that is protected by legal privilege or by doctor-patient confidentiality, data that includes the content of individuals' non-public communications)? \\
        \answerNo{}
        \item Does the dataset contain data that, if viewed directly, might be offensive, insulting, threatening, or might otherwise cause anxiety? \\
        \answerNo{}
        \item Does the dataset relate to people? \\
        \answerNo{}
        \item Does the dataset identify any subpopulations (e.g., by age, gender)? \\
        \answerNo{}
        \item Is it possible to identify individuals (i.e., one or more natural persons), either directly or indirectly (i.e., in combination with other data) from the dataset? \\
        \answerNo{}
        \item Does the dataset contain data that might be considered sensitive in any way (e.g., data that reveals racial or ethnic origins, sexual orientations, religious beliefs, political opinions or union memberships, or locations; financial or health data; biometric or genetic data; forms of government identification, such as social security numbers; criminal history)?  \\
        \answerNo{}
        \item Any other comments? \\
        \answerNo{}
    \end{enumerate}
    \item Collection Process
    \begin{enumerate}
        \item How was the data associated with each instance acquired? \\
        For each instance, we capture 24 profile images. For each testing image, we manually annotate bounding boxes on the instances of interest as the ground-truth.
        \item What mechanisms or procedures were used to collect the data (e.g., hardware apparatus or sensor, manual human curation, software program, software API)? \\
        We use a single Leica camera (embedded in a cellphone) to capture profile images and testing scene images. We use GrabCut to obtain foreground masks on the profile images. We manually draw bounding boxes on testing images as the ground-truth. 
        \item If the dataset is a sample from a larger set, what was the sampling strategy (e.g., deterministic, probabilistic with specific sampling probabilities)? \\
        \answerNA{}
        \item Who was involved in the data collection process (e.g., students, crowdworkers, contractors), and how were they compensated (e.g., how much were crowdworkers paid)? \\
        Only authors are involved in the data collection.
        \item Over what timeframe was the data collected? \\
        All images were collected between October 2022 to March 2023. 
        \item Were any ethical review processes conducted (e.g., by an institutional review board)? \\
        \answerNo{No ethical review processes were conducted with respect to the collection and annotation of this data.}
        \item Does the dataset relate to people? \\
        \answerNo{}
        \item Did you collect the data from the individuals in question directly, or obtain it via third parties or other sources (e.g., websites)? \\
        We collected the data by ourselves. 
        \item Were the individuals in question notified about the data collection? \\
        \answerYes{} 
        \item Did the individuals in question consent to the collection and use of their data? \\
        \answerYes{} 
        \item If consent was obtained, were the consenting individuals provided with a mechanism to revoke their consent in the future or for certain uses? \\
        \answerNo{} 
        \item Has an analysis of the potential impact of the dataset and its use on data subjects (e.g., a data protection impact analysis) been conducted? \\
        \answerNA{}
        \item Any other comments? \\
        \answerNo{}
    \end{enumerate}
    \item Preprocessing, Cleaning and Labeling
    \begin{enumerate}
        \item Was any preprocessing/cleaning/labeling of the data done (e.g., discretization or bucketing, tokenization, part-of-speech tagging, SIFT feature extraction, removal of instances, processing of missing values)? \\
        \answerNo{The only labeling activity is annotating bounding boxes for instances on testing images. Further annotation cleaning is not needed.}
        \item Was the "raw" data saved in addition to the preprocessed/cleaned/labeled data (e.g., to support unanticipated future uses)? \\
        \answerYes{We release the raw data on the website of this work}.
        \item Is the software used to preprocess/clean/label the instances available? \\
        \answerYes{We use open-source software publicly available. We cite them already in the main paper or supplement.}
        \item Any other comments? \\
        \answerNo{}
    \end{enumerate}
    \item Uses
    \begin{enumerate}
        \item Has the dataset been used for any tasks already? \\
        \answerNo{}
        \item Is there a repository that links to any or all papers or systems that use the dataset? \\
        \answerNo{We will add future papers that use this dataset in the website of this work.}
        \item What (other) tasks could the dataset be used for? \\
        The dataset can also be used to study small object detection, 3D reconstruction from sparse views, real-time object detection, etc.
        \item Is there anything about the composition of the dataset or the way it was collected and preprocessed/cleaned/labeled that might impact future uses? \\
        \answerYes{Capturing high-resolution images is cheap but requires more computation resources in subsequent tasks. This might impact future work related to how to capture high-resolution images, how to trade off performance and resolution, etc.}
        \item Are there tasks for which the dataset should not be used? \\
        The usage of this dataset should be limited to the scope of object instance detection.
        \item Any other comments? \\
        \answerNo{}
    \end{enumerate}
    \item Distribution
    \begin{enumerate}
        \item Will the dataset be distributed to third parties outside of the entity (e.g., company, institution, organization) on behalf of which the dataset was created? \\
        \answerYes{We expect other websites re-distribute our dataset.}
        \item How will the dataset be distributed (e.g., tarball on website, API, GitHub)? \\
        The dataset could be accessed on a GitHub webpage.
        \item When will the dataset be distributed? \\
        The dataset will be released to the public upon acceptance of this paper. We provide some demo data and visualizations for the review process.
        \item Will the dataset be distributed under a copyright or other intellectual property (IP) license, and/or under applicable terms of use (ToU)? \\
        We release our benchmark under MIT license.
        \item Have any third parties imposed IP-based or other restrictions on the data associated with the instances? \\
        \answerNo{}
        \item Do any export controls or other regulatory restrictions apply to the dataset or to individual instances? \\
        \answerNo{}
        \item Any other comments? \\
        \answerNo{}
    \end{enumerate}
    \item Maintenance
    \begin{enumerate}
        \item Who is supporting/hosting/maintaining the dataset? \\
        Qianqian Shen will be responsible for maintaining the dataset. 
        \item How can the owner/curator/manager of the dataset be contacted (e.g., email address)? \\
        E-mail addresses are at the top of the paper.
        \item Is there an erratum? \\
        \answerNo{When errors are discerned, we will announce erratum on our website}.
        \item Will the dataset be updated (e.g., to correct labeling errors, add new instances, delete instances')? \\
        \answerYes{We hope to expand this dataset with more instances and testing images.}
        \item If the dataset relates to people, are there applicable limits on the retention of the data associated with the instances (e.g., were individuals in question told that their data would be retained for a fixed period of time and then deleted)? \\
        \answerNA{}
        \item Will older versions of the dataset continue to be supported/hosted/maintained? \\
        \answerYes{}
        \item If others want to extend/augment/build on/contribute to the dataset, is there a mechanism for them to do so? \\
        \answerNo{}
        \item Any other comments? \\
        \answerNo{}
    \end{enumerate}
\end{enumerate}

\end{document}